\documentclass[final]{svjour3}                     % onecolumn (standard format)
\smartqed  % flush right qed marks, e.g. at end of proof
\usepackage{graphicx}
\usepackage{floatrow}
\usepackage{bm}
\usepackage{multirow}
\usepackage{mathrsfs,amsmath}
\usepackage{threeparttable}
\usepackage{array}
\usepackage{graphicx}%
\usepackage{changepage}
\usepackage{setspace}
\usepackage{pdfpages}
\usepackage{pdflscape}
\usepackage[numbers,sort&compress]{natbib}
\usepackage{graphicx}
\usepackage{multirow}
\usepackage[figuresright]{rotating}
\usepackage{times}
\newfloatcommand{capbtabbox}{table}[\capbottom][\FBwidth]
\usepackage{caption}
\usepackage{algorithm,algorithmic}

\usepackage{makecell}
\usepackage{multirow}
\usepackage{appendix}
\usepackage{graphicx}
\usepackage{subfigure}
\usepackage{dsfont,bm}
\usepackage{threeparttable}
\usepackage{setspace}
\usepackage{lscape}
\usepackage{lscape}
\usepackage [latin1]{inputenc}

\newcommand{\tabincell}[2]{\begin{tabular}{@{}#1@{}}#2\end{tabular}}
 \journalname{Data Mining and Knowledge Discovery}
\begin{document}

\title{Jointly Dynamic Topic Model for Recognition of Lead-lag Relationship in Two Text Corpora
%\thanks{Correspondence to Feifei Wang}
}

\author{Yandi Zhu         \and
        Xiaoling Lu \and
        Jingya Hong \and
        Feifei Wang
}

\institute{Yandi Zhu \at
              School of Economics, Peking University, Beijing, China
            \and
           Jingya Hong \at
           School of Statistics, Renmin University of China, Beijing, China
           \and
           Xiaoling Lu \and Feifei Wang \at
           Center for Applied Statistics, Renmin University of China, Beijing, China \\
           School of Statistics, Renmin University of China, Beijing, China\\
           %\email{{\color{magenta}XXXX}}           %  \\          %  \\
           \\
           Xiaoling Lu is the co-first author.\\
              Correspondence to: Feifei Wang \email{feifei.wang@ruc.edu.cn}       %  \\
}

\date{Received: date / Accepted: date}

\maketitle

\begin{abstract}
Topic evolution modeling has received significant attentions in recent decades. Although various topic evolution models have been proposed, most studies focus on the single document corpus. However in practice, we can easily access data from multiple sources and also observe relationships between them. Then it is of great interest to recognize the relationship between multiple text corpora and further utilize this relationship to improve topic modeling. In this work, we focus on a special type of relationship between two text corpora, which we define as the ``lead-lag relationship". This relationship characterizes the phenomenon that one text corpus would influence the topics to be discussed in the other text corpus in the future. To discover the lead-lag relationship, we propose a jointly dynamic topic model and also develop an embedding extension to address the modeling problem of large-scale text corpus. With the recognized lead-lag relationship, the similarities of the two text corpora can be figured out and the quality of topic learning in both corpora can be improved. We numerically investigate the performance of the jointly dynamic topic modeling approach using synthetic data. Finally, we apply the proposed model on two text corpora consisting of statistical papers and the graduation theses. Results show the proposed model can well recognize the lead-lag relationship between the two corpora, and the specific and shared topic patterns in the two corpora are also discovered.

\end{abstract}

\section{Introduction}
Dynamic documents have become commonly encountered in our daily life. Facing the increasing accumulation of text streams, how to quickly grasp the pattern of topic changes underlying the text corpus becomes a problem of great interest. To address this issue, dynamic topic models, a suite of three-level hierarchical Bayesian models originated from the latent Dirichlet allocation (LDA) \citep{blei2003latent}, have been widely used to reveal the thematic structure in large dynamic document corpus. Typical examples include: the dynamic topic model (DTM) \citep{Blei2006Dynamic}, which assumes an evolution pattern for hyperparameter $\beta$s to model topic changes; the multiscale topic tomography model (MTTM) \citep{Nallapati2007Multiscale}, which assumes both data generation process and parameter generative process over time; the continuous time dynamic topic model (cDTM) \citep{Wang2008Continuous}, which applies Brownian motion for detecting topic evolution; and other recent works \citep{2011Trend,Chae2012Spatiotemporal,2013non,Zhou:2014:EDO:2628707.2628786,2015He,2019online}.

\begin{figure}[!h]
	\centering
	\includegraphics[width=0.9\linewidth]{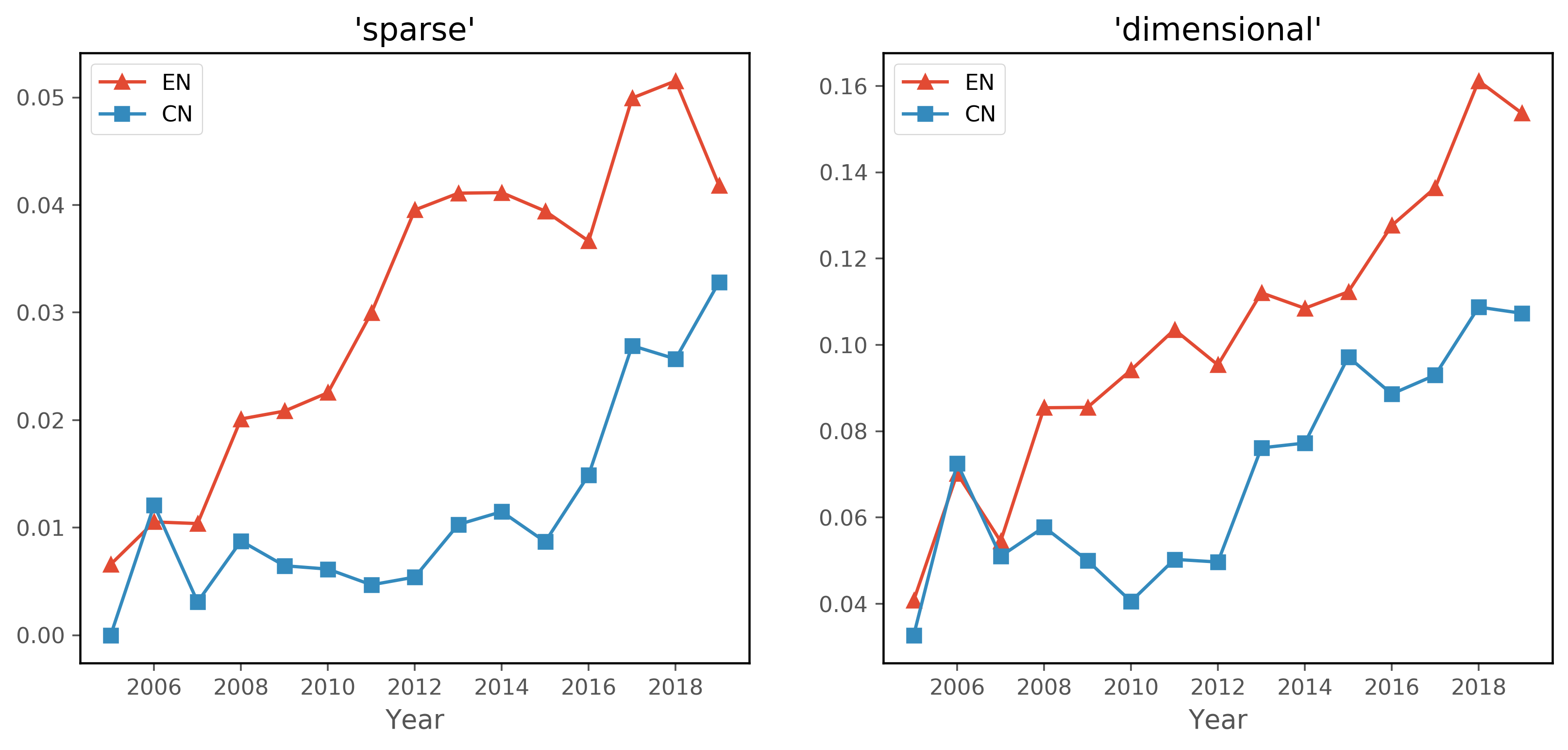}
	\caption{The time varying frequencies of ``sparse" and ``dimensional" in the journal corpus EN and the graduation theses corpus CN. As shown, the journal corpus \emph{leads} the trend while the thesis corpus \emph{is lagged} almost three years. Then the journal corpus serves as the \emph{leading} corpus and the thesis corpus serves as the \emph{lagged} corpus. There exists a lead-lag relationship embedded between the two corpora.}
	\label{fig:Words_prop}
\end{figure}

Although various works have been proposed for topic evolution modeling, most studies focus on the single document corpus and try to recognize the pattern of topic changes reflected by its own. With the accumulation of text documents in all fields, we can easily access data from multiple sources and also observe relationships between them. For example, in academic fields, research scholars often discuss cutting-edge research issues in academic papers. These fresh ideas lead the academic trends and thus exert influences on the choice of thesis topics for graduate students. To verify this idea, we take the field of \emph{statistics} as an example. We collected two text corpora: the collection of papers published on top statistical journals, which is denoted as EN; and the collection of Chinese graduation theses majored in statistics, which is denoted as CN. The detailed data description is present in Section 5.1. We then selected two words, ``sparse" and ``dimensional", which are classic topics in statistics. Figure \ref{fig:Words_prop} compares the time varying frequencies of the two words in both corpora. As shown, the journal corpus EN \emph{leads} the trend while the thesis corpus CN \emph{is lagged} almost three years. We define the journal corpus as the \emph{leading corpus} while the thesis corpus as the \emph{lagged corpus}. The relationship between these two corpora is then defined as the \emph{lead-lag relationship}. This relationship characterizes the phenomenon that the lagged corpus follows the topics which have been discussed by the leading corpus before.

Except for academic documents, the lead-lag relationship can be found in a variety of fields. For example, in stock investment, the institutional research reports (served as leading corpus) could influence the online discussions of shareholders (served as lagged corpus). In news media, the breaking news posted by mainstream media (served as leading corpus) could influence the community discussions in social network platforms (served as lagged corpus). Recognition of the lead-lag relationship in two text corpora can help better understand the relationship of the two corpora. Essentially, the lead-lag relationship of the two corpora is the \emph{causality} in time series. That is, the topics discussed in the leading corpus could \emph{cause} the changes of topics in the lagged corpus. Consequently, with the recognized lead-lag relationship, researchers can better understand the dynamic changes of two text corpora. In addition, the topics to be discussed in the lagged corpus can be forecast in advance. From the perspective of topic modeling, as more information is considered (i.e., the two corpora share some same information), the quality of topics shared by these two corpora can be improved. Therefore, it is of great interest to develop algorithms to exploit the lead-lag relationship between two corpora.

To this end, we propose a jointly dynamic topic model (JDTM) to investigate the lead-lag relationship between two text corpora. Specifically, we assume there are three types of topics: (1) the shared topics, which are represented by both leading corpus and lagged corpus; (2) the lead-specific topics, which are only represented by the leading corpus, and (3) the lag-specific topics, which are only represented by the lagged corpus. It is notable that, although the shared topics are represented by both corpora, the representing time is different. Assume the leading corpus has an $l$-period topic influence on the lagged corpus. Then the leading corpus at time $t$ and the lagged corpus at time $t+l$ would represent the same shared topics. We then follow the dynamic topic model \citep{Blei2006Dynamic} to characterize the evolution patterns of the three types of topics in the two corpora, but with the lead-lag relationship embedded in the shared topics. In addition, to handle challenges imposed by large-scale textual datasets, we apply the embedding techniques and develop a jointly dynamic embedding topic model (JDETM).

To summarize, our main contributions are present as follows:
\begin{itemize}
    \item We focus on the topic modeling problem of two text corpora, and define a ``lead-lag" relationship to characterize the phenomenon that one text corpus would influence the topics to be discussed in the other text corpus in the future.
	\item We propose a jointly dynamic topic model to investigate the lead-lag relationship between two text corpora, which to our best knowledge has never been considered in topic modeling. The proposed model can help recognize the specific and shared patterns in the two corpora.
    \item To handle large-scale textual datasets, an embedding version of JDTM is further developed. By modeling topics in the embedding spaces, both topic learning and document representations can be enhanced. It is also notable that, the jointly dynamic modeling approach can be easily extended to other dynamic topic models to recognize more relationships between two corpora.
	\item To evaluate the performance of JDTM and JDETM, we conduct various experiments on the synthetic data. In addition, we apply the proposed model on the academic documents with statistical papers as the leading corpus and the graduation theses as the lagged corpus. Results show the proposed model can well recognize the lead-lag relationship between the two corpora.
\end{itemize}

The rest of the paper is organized as follows. In Section 2, we briefly review the related works of topic models for dynamic documents. In Section 3, we introduce the model description and estimation algorithms of JDTM and JDETM in details. In Section 4, we conduct a variety of experiments on synthetic datasets to evaluate the performance of JDTM and JDETM. In Section 5, we apply the proposed model on the two corpora of statistical papers and graduation theses to explore their lead-lag relationship. Section 6 presents the conclusions and a brief discussion.

\section{Related Work}

To model topic evolutions or storylines for dynamic documents, a variety of topic models have been proposed \citep{Wang2008Continuous,Chae2012Spatiotemporal,Zhou:2014:EDO:2628707.2628786,Pozdnoukhov2011Space,Topics2006wang,Vavliakis12eventdetection}.
According to \citep{2017Topic}, the topic evolution models can be roughly classified into three categories according to the type of time variable. Topic evolution models for discrete time are the most common category. For example, the dynamic topic model \citep[DTM,][]{Blei2006Dynamic} assumes the topics in slice $t$ are evolved from the corresponding topics in slice $t-1$. By this way, the changes of topics in dynamic documents can be captured. Other important models in this category include the multiscale topic tomography model \citep[MTTM,][]{Nallapati2007Multiscale}, temporal Dirichlet process mixture model \citep[TDPM,][]{2008Dynamic}, infinite dynamic topic model \citep[iDTM,][]{2010Timeline}, and so on. The second category is the topic evolution model for continuous time. A typical example is the continuous time dynamic topic model \citep[cDTM,][]{Wang2008Continuous}, which uses Brownian motion to model continuous-time topic drift. Other extensions of this category include \citep{2011Trend,2013non,Topics2006wang}. The last category is the online topic evolution model, which attracts more attentions in the recent decade. For example, \cite{2008online} extends the original LDA model into an online fashion and a solution based on the empirical Bayes is developed for estimation. Other works include \cite{2015He,2019online,2014online}. Although various works have been proposed for topic evolution modeling, the above literature mainly focuses on the single corpus. In this work, we focus on two text corpora and try to investigate the embedded lead-lag relationship between them.

Some recent works try to improve topic modelling performance using word embeddings and deep neural networks, and have good performance in modeling very large text corpora. For example, \citep{rudolph2018dynamic} develops dynamic embeddings, which build on exponential family to capture how the meanings of words change over time. \citep{dieng2020topic} models topics in embedding spaces and defines the categorical distribution through the inner product between word embeddings and topic embeddings. Another typical example is the dynamic embedding topic model (DETM) \citep{dieng2019the}, which learns smooth topic evolutions by defining a random walk prior over the embedding representations of the topics and uses structured amortized variational inference with a recurrent neural network. More works enhancing topic models by word embeddings or word vectors can be seen in \citep{2019Discovering,2019A,2021Jointly}. In this work, we also borrow ideas from DETM to develop a jointly dynamic embedding topic model to recognize the lead-lag relationship between two large-scale text corpora.

The lead-lag relationship we discovered in this work is similar to the temporal dependence in time series analysis. Therefore, our work is also closely related to the time series methods for recognition of temporal dependence underlying dynamic systems. To discovery the relationship between two time series, one commonly used tool is the \emph{lagged cross-correlation function}, which calculates the correlation coefficients between two time series under different time shifts \citep{2008Time}. The Granger causality test further identifies dependence structure between two time series through predictability using parametric autoregressive models \citep{granger1969investigating,ashley1980advertising}. Nonparametric models based on optimal path searching are also proposed, which can alleviate the limitation of data volume in Granger causality test \citep{sornette2005non,meng2017symmetric}. However, all these methods cannot fully identify the ``causality" relationship between two time series. To address this issue, state space reconstruction methods have been developed, which can better address the nonlinear state-dependent couplings \citep{sugihara2012detecting,ye2015distinguishing}. For example, the convergent cross mapping (CCM) method \citep{sugihara2012detecting} tests for causation by measuring the extent to which the historical record of one time series can reliably estimate the states of the other one based on nonlinear state space reconstruction. These methods provide a generalized approach to detect causality in complex systems. See \citep{runge2019inferring} for more discussions about the causal discovery for time series data. In this work, we would apply the CCM method to verify the underlying causal structure discovered by our method.

\section{The Jointly Dynamic Topic Model}

\subsection{Model description}

Assume we have two text corpora, i.e., the leading corpus $\mathcal{C}_{lead}$ and the lagged corpus $\mathcal{C}_{lag}$. Both corpora are collected for time $t=1,2,...,T$. Accordingly, the leading corpus and lagged corpus can be split into small datasets according to time. That is, $\mathcal{C}_{lead} = \{\mathcal{C}_{lead,1},...,\mathcal{C}_{lead,T}\}$, and $\mathcal{C}_{lag} = \{\mathcal{C}_{lag,1},...,\mathcal{C}_{lag,T}\}$. In the time slice $t$, assume $\mathcal{C}_{lead,t}$ contains a number of $D_{lead,t}$ documents, while $\mathcal{C}_{lag,t}$ contains a number of $D_{lag,t}$ documents. For each single document $d$ in the two corpora, it contains $N_d$ words, which can be represented by $\textbf{w}_d=\{w_{d,1},w_{d,2},...,w_{d,N_d}\}$.

Given the two text corpora have some content similarities, we assume three types of topics underlying the two corpora. First, assume there are $K$ shared topics, which are represented by both the leading corpus and lagged corpus. Denote the $k$-th shared topic at time $t$ by $\alpha_{t,k}$, which has a probability distribution over the vocabulary of $V$ words. Second, assume there are $J$ lead-specific topics, which are only represented by the leading corpus. Denote the $j$-th lead-specific topic at time $t$ by $\beta_{t,j}$. Last, assume there are $H$ lag-specific topics, which are only represented by the lagged corpus. Denote the $h$-th lag-specific topic at time $t$ by $\gamma_{t,h}$. Both $\beta_{t,j}$ and $\gamma_{t,h}$ have probability distributions over the vocabulary of $V$ words. With the three types of topics, the leading corpus and lagged corpus not only share some similarities, but also have their own content characteristics.

The lead-lag relationship is represented by the shared topics, but with an $l$-period time interval between the leading corpus and lagged corpus. Specifically, for document $d$ in the leading corpus at time $t$, it has a distribution $\theta_d$ over $K$ shared topics and $J$ lead-specific topics. Therefore, $\theta_d$ is a $(K+J)$-dimensional probability vector. Each word in document $d$ can represent either the shared topics at time $t$ or the lead-specific topics at time $t$. For the document $d'$ in the lagged corpus at time $t$, it has a distribution $\phi_{d'}$ over $K$ shared topics and $H$ lag-specific topics. Thus $\phi_{d'}$ is a $(K+H)$-dimensional probability vector. Given the shared topics in the lagged corpus are $l$-period lagged behind those in the leading corpus, each word in document $d'$ can represent the lag-specific topics at time $t$ or the shared topics at time $t+l$. In other words, the shared topics are totally determined by the leading corpus. Once the shared topics are decided by the leading corpus at time $t$, one can forecast the topics underlying the lagged corpus at time $t+l$.

\begin{figure}[h]
	\centering
	\includegraphics[width=0.95\linewidth]{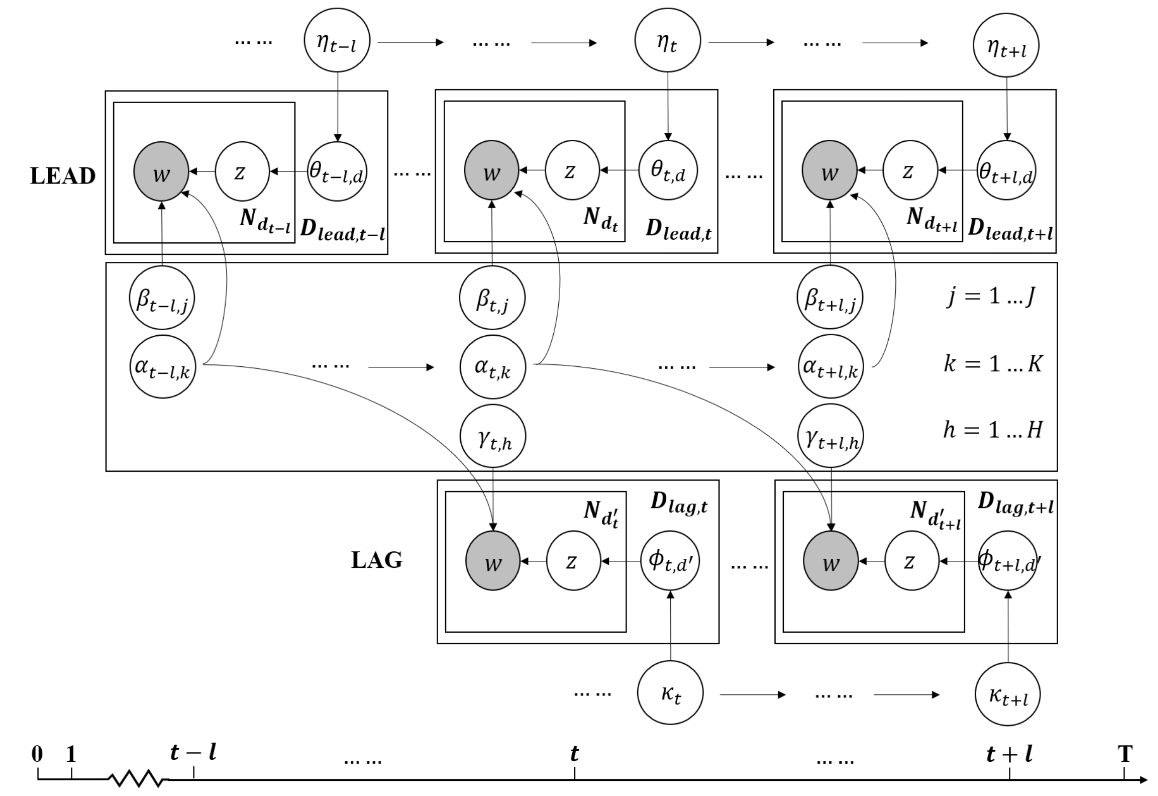}
	\caption{The generative process of the jointly dynamic topic model with the lagged time period equal to $l$. For documents in the leading corpus at time $t$, they represent
the lead-specific topics $\{\beta_{t,j}\}$s and the shared topics $\{\alpha_{t,k}\}$s. For documents in the lagged corpus at time $t$, they represent the lag-specific topics $\{\gamma_{t,j}\}$s and the shared topics $\{\alpha_{t-l,j}\}$s.}
	\label{fig:ourmodel}
\end{figure}

Furthermore, we follow the dynamic topic model \citep{Blei2006Dynamic} to characterize the evolution pattern of each topic. Specifically, we model the natural parameters of each topic using a Gaussian process and then map the emitted values to the simplex by softmax transformation. Together with the above assumptions, the generative process of documents is present below:
\begin{enumerate}
	\item For time slice $t = 1, 2, ..., T$:
	\begin{enumerate}
		\item For $1 \leq k \leq K$, draw shared topics $\alpha_{t,k}|\alpha_{t-1,k} \sim N(\alpha_{t-1,k},\sigma^2_kI)$
		\item For $1 \leq j \leq J$, draw lead-specific topics $\beta_{t,j}|\beta_{t-1,j}\sim N(\beta_{t-1,j},\sigma^2_jI)$
		\item For $1 \leq h \leq H$, draw lag-specific topics $\gamma_{t,h}|\gamma_{t-1,h}\sim N(\gamma_{t-1,h},\sigma^2_hI)$
	\end{enumerate}
	\item For the leading corpus:
	\begin{enumerate}
		\item Draw the mean of topic proportions $\eta_t|\eta_{t-1} \sim N(\eta_{t-1},\delta^2_f I)$
		\item For document $d$ whose time stamp is $t_d$:
		\begin{enumerate}
			\item Draw topic proportions $\theta_d \sim N(\eta_{t_d},\varpi_f^2 I)$
			\item For each word $n$ in this document:
			\begin{enumerate}
				\item Draw topic assignment $z_{dn}\sim Multi(\text{softmax}(\theta_d))$
				\item If $z_{dn}$ represents a shared topic, then draw word

$w_{dn} \sim Multi(\text{softmax}(\alpha_{t_d,z_{dn}}))$
				\item If  $z_{dn}$ represents a lead-specific topic, then draw word

$w_{dn} \sim Multi(\text{softmax}(\beta_{t_d,z_{dn}}))$
			\end{enumerate}
		\end{enumerate}
	\end{enumerate}
	\item For the lagged corpus:
	\begin{enumerate}
		\item Draw the mean of topic proportions $\kappa_t|\kappa_{t-1} \sim N(\kappa_{t-1},\delta^2_l I)$
		\item For document $d'$ whose time stamp is $t_{d'}$:
		\begin{enumerate}
			\item Draw topic proportions $\phi_{d'} \sim N(\kappa_{t_{d'}},\varpi_l^2 I)$
			\item For each word $n$ in this document:
			\begin{enumerate}
				\item Draw topic assignment $z_{{d'}n} \sim Multi(\text{softmax}(\phi_{d'}))$
				\item If $z_{{d'}n}$ represents a shared topic, then draw word

$w_{{d'}n} \sim Multi(\text{softmax}(\alpha_{t_{d'}-l,z_{{d'}n}}))$
				\item If $z_{{d'}n}$ represents a lag-specific topic, then draw word

$w_{{d'}n} \sim Multi(\text{softmax}(\gamma_{t_{d'},z_{{d'}n}}))$
			\end{enumerate}
		\end{enumerate}
	\end{enumerate}
\end{enumerate}
In the above generative process, the $\text{softmax}$ function maps the mean parameters to the multinomial natural parameters. Denote $x=(x_1,,,x_g)^{\top}$ be an arbitrary $g$-dimensional vector. Then $\text{softmax}(x)=\exp(x)/\sum_{i=1}^{g} \exp(x_i)$. For illustration purpose, the generative process of JDTM is summarized in Figure \ref{fig:ourmodel}.

\subsection{Model inference}

We use the method of variational inference for model estimation. Denote the collection of word probabilities for the three types of topics as $\bm{\alpha}=\{\alpha_{t,k}:1 \leq t \leq T, 1 \leq k \leq K\}$, $\bm{\beta}=\{\beta_{t,k}:1 \leq t \leq T, 1 \leq k \leq K\}$, and $\bm{\gamma}=\{\gamma_{t,k}:1 \leq t \leq T, 1 \leq k \leq K\}$. Denote the topic proportions for each document in the two corpora as $\bm{\theta}=\{\theta_{d}:d \in \mathcal{C}_{lead}\}$ and $\bm{\phi}=\{\phi_{d'}:d' \in \mathcal{C}_{lag}\}$. Denote the collection of topic assignments for all words in the two corpora as $\bm{z}$. Following \cite{Blei2006Dynamic}, we consider $\{\eta_t\}$s and $\{\kappa_t\}$s as hyperparameters, which are specified in advance. Then the collection of all hyperparameters is $\Lambda=\{\sigma_k^2,\sigma_j^2,\sigma_h^2,\delta_f^2,\delta_l^2,\varpi_f^2,\varpi_l^2,\eta_t,\kappa_t\}$. Given the two text corpora $\mathcal{C}_{lead}$ and $\mathcal{C}_{lag}$, as well as the hyperparameters $\Lambda$, the posterior distribution is derived as follows
\begin{equation}
\label{eq:post}
	p=p(\bm{\alpha},\bm{\beta},\bm{\gamma},\bm{z},\bm{\theta},\bm{\phi}|\mathcal{C}_{lead}, \mathcal{C}_{lag},\Lambda),	
\end{equation}
By simple calculations, we find the posterior in \eqref{eq:post} is intractable. Thus, we employ variational inference \citep{Jordan1999An} for model estimation. The idea of variational inference is to posit a distribution family with free parameters (called variational parameters) over the latent variables and then try to find the optimal parameters that can minimize the Kullback-Leibler (KL) divergence to the true posterior.

In our model, we specify the variational distribution to the true posterior in \eqref{eq:post} as follows. First, the document-level latent variables follow the same specification as \cite{blei2003latent}. Specifically, for each word $w_{t,d,n}$ from the document $d$ in the leading corpus, its hidden topic assignment $z_{t,d,n}$ is specified by a multinomial distribution with parameter $\lambda_{t,d,n}$. The vector of topic proportions $\theta_{t,d}$ is endowed with a free Dirichlet parameter $\tau_{t,d}$. Recall that the leading corpus only represents the $K$ shared topics and $J$ lead-specific topics. Therefore,  $\lambda_{t,d,n}$ and $\tau_{t,d}$ are $(K+J)$-dimensional vectors. Similar specifications are assumed for the lagged corpus, but the variational parameters are all $(K+H)$-dimensional vectors. Second, we assume the specification of topic trajectories following \cite{Blei2006Dynamic}. Specifically, for each type of topics, we posit a dynamic model with Gaussian ``variational observations". For example, for the $k$-th shared topic, we assume its variational distribution is {$q(\alpha_{1,k},...,\alpha_{T,k}|\hat{\alpha}_{1,k},...,\hat{\alpha}_{T,k})$}. Denote the collection of all variational parameters as $\Gamma$. Given the two corpora are generated independently, the variational distribution is specified as follows
\begin{equation}
\begin{aligned}
q= \ &q(\bm{\alpha},\bm{\beta},\bm{\gamma},\bm{z},\bm{\theta},\bm{\phi}| \mathcal{C}_{lead}, \mathcal{C}_{lag},\Gamma)\\
= \ & \prod_{k=1}^K q(\alpha_{1,k},...,\alpha_{T,k}|\hat{\alpha}_{1,k},...,\hat{\alpha}_{T,k})
\prod_{j=1}^J q(\beta_{1,j},...,\beta_{T,j}|\hat{\beta}_{1,j},...,\hat{\beta}_{T,j}) \\
\times \ & \prod_{h=1}^H q(\gamma_{1,h},...,\gamma_{T,h}|\hat{\gamma}_{1,h},...,\hat{\gamma}_{T,h})
\prod_{t=1}^T\prod_{d=1}^{D_{leag,t}}\{q(\theta_{t,d}|\tau_{t,d})\prod_{n=1}^{N_{d}} q(z_{t,d,n}|\lambda_{t,d,n})\} \\
\times \ & \prod_{t=1}^T\prod_{d'=1}^{D_{lag,t}}\{q(\phi_{t,d'}|\tau_{t,d'})\prod_{n=1}^{N_{d'}} q(z_{t,d',n}|\lambda_{t,d',n})\}. \nonumber
\end{aligned}
\end{equation}

The goal is to solve the optimization problem, i.e, finding the optimal variational parameters that can minimize $KL(q||p)$. This is equivalent to maximizing the evidence lower bound (ELBO),
\begin{equation}
\label{jdtm:elbo}
\begin{aligned}
\mathscr{L}(q)
=&\text{E}_q\left[\log  p-\log q \right]\\
=&\sum_{t=1}^{T}\text{E}_q[\log p(\alpha_t|\alpha_{t-1})+\log p(\beta_t|\beta_{t-1})+\log p(\gamma_t|\gamma_{t-1})] \\
+&\sum_{t=1}^{T}\sum_{d=1}^{D_{lead,t}}\text{E}_q[\log p(\theta_{t,d}|\eta_t)]+\sum_{t=1}^{T}\sum_{d'=1}^{D_{lag,t}}\text{E}_q[\log p(\phi_{t,d'}|\kappa_t)] \\
+&\sum_{t=1}^{T}\sum_{d=1}^{D_{lead,t}}\sum_{n=1}^{N_d}\text{E}_q[\log p(z_n|\theta_{d})+\log p(w_n|z_n,\alpha_{t},\beta_{t})] \\
+&\sum_{t=1}^{T}\sum_{d'=1}^{D_{lag,t}}\sum_{n=1}^{N_{d'}}\text{E}_q[\log p(z_n|\phi_{d'})+\log p(w_n|z_n,\alpha_{t-l},\gamma_{t})]
+H(q).
\end{aligned}
\end{equation}
We optimize the ELBO by using the coordinate ascent method, which iteratively optimizes each parameter in the variational distribution $q$. These updates are repeated until \eqref{jdtm:elbo} has converged. Below, we discuss the updating formula for each variational parameter.

\textbf{Topic trajectories.} The updates for the topic-level variational parameters do not have a closed form. Following \cite{Blei2006Dynamic}, we adopt a conjugate gradient method based on Kalman filter \citep{kalman1960a}. In general, the updates for the three types of topics are similar. Therefore, we take the update of shared topics as an example. Denote $\hat{\alpha}_{t,k,v}$ as the probability of the $k$-th shared topic on the $v$-th word at time $t$. Then the gradient with respect to $\hat{\alpha}_{t,k,v}$ is,
\begin{equation}
\label{jdtm:update:topic}
\begin{aligned}
	\frac{\partial \mathscr{L}}{\partial \hat{\alpha}_{t,k,v}} & = -\frac{1}{\sigma^2_k}\sum_{s=1}^{T}(\widetilde{m}_{s,k,v}-\widetilde{m}_{s-1,k,v})\bigg(\frac{\partial \widetilde{m}_{s,k,v}}{\partial\hat{\alpha}_{t,k,v}}-\frac{\partial \widetilde{m}_{s-1,k,v}}{\partial
		\widetilde{m}_{s-1,k,v}}\bigg)\\
&+\sum_{s=1}^T\big(n_{s,k,v}-n_{s,k}\hat{\zeta}_{s,k}^{-1}\exp(\widetilde{m}_{s,k,v}+\widetilde{V}_{s,k,v}/2)\big)\frac{\partial\widetilde{m}_{s,k,v}}{\partial\hat{\alpha}_{t,k,v}}
\end{aligned}
\end{equation}
where $\widetilde{m}_{s,k,v}$ and $\widetilde{V}_{s,k,v}$ are the backward mean and variance, $\hat{\zeta}_{s,k}$ is an additional variational parameter. All the notations are defined as the same as \cite{Blei2006Dynamic}. The only difference among the three types of topics lies in the calculation of $n_{t,k,v}$, which is the number of word $v$ representing the $k$-th topic at time $t$. Specifically, for the shared topic $1 \leq k \leq K$, $n_{t,k,v}$ is calculated as follows
\begin{equation}
n_{t,k,v}=\sum_{d=1}^{D_{lead,t}}\sum_{n=1}^{N_{d}}I[w_{t,d,n}=v]\lambda_{t,d,n,k}
+\sum_{d'=1}^{D_{lag,t}}\sum_{n=1}^{N_{d'}}I[w_{t,d',n}=v]\lambda_{t,d',n,k},\nonumber
\end{equation}
where $I(\cdot)$ is the indicator function. For the lead-specific topic $1 \leq j \leq J$ and the lag-specific topic  $1 \leq h \leq H$, the calculation of $n_{t,j,v}$ and $n_{t,h,v}$ are shown below:
\begin{equation}
\begin{split}
n_{t,j,v}&=\sum_{d=1}^{D_{lead,t}}\sum_{n=1}^{N_{d}}I[w_{t,d,n}=v]\lambda_{t,d,n,j}\\
n_{t,h,v}&=\sum_{d=1}^{D_{lag,t}}\sum_{n=1}^{N_{d'}}I[w_{t,d',n}=v]\lambda_{t,d',n,h}.\nonumber
\end{split}
\end{equation}

\textbf{Topic proportions and topic assignments.} The updates for document-level variational parameters have a closed form as shown in \cite{blei2003latent}. Specifically, the updates for the variational Dirichlet parameters for topic proportions are shown below:
\begin{equation}
\label{jdtm:tau:d}
	\tau_{t,d} = \eta_{t} + \sum_d^{N_d}\lambda_{t,d,n}, \mbox{~~}
	\tau_{t,d'} = \kappa_{t} + \sum_{d'}^{N_d'}\lambda_{t,d,n},
\end{equation}
where $d$ and $d'$ represent an arbitrary document in the leading corpus and lagged corpus, respectively.

Given the variational topic distributions, we update the variational parameters for topic assignments as follows:
\begin{equation}
\begin{split}
\label{jdtm:lambda:d}
	\log \lambda_{t,d,n,k} &\propto \widetilde{m}_{t,k,n} - \log \zeta_{t,k}+ \Psi(\tau_{t,d,k}) \\
	\log \lambda_{t,d',n,k} &\propto \widetilde{m}_{t,k,n} - \log \zeta_{t,k} + \Psi(\tau_{t,d',k}),
\end{split}
\end{equation}
where $\Psi$ is the diagamma function. In each round of recursion, we first update the document-level parameters and then fit the variational observations from expected counts under those document-level variational distributions. Algorithm \ref{alg1} summarizes the estimation procedure of JDTM.
\begin{algorithm}[h]
	\caption{Model Estimation Procedure for JDTM}
	\label{alg1}
	\begin{algorithmic}[1]
		\REQUIRE $\mathcal{C}_{lead}$, $\mathcal{C}_{lag}$, $\Lambda$, $\Gamma$, $K$, $J$, $H$, and $l$.
		\STATE Initialize all variational parameters
		\WHILE{the ELBO has not converged,}
		\FOR{$t$ in $1,2,...,T$}
		\FOR{each document $d$ in the leading corpus $\mathcal{C}_{lead,t}$}
		\FOR{each word $n$ in the document}
		\STATE Update topic assignments according to \eqref{jdtm:lambda:d}
		\ENDFOR
		\STATE Update topic proportions according to \eqref{jdtm:tau:d}
		\ENDFOR
		\FOR{each document $d'$ in the lagged corpus $\mathcal{C}_{lag,t}$}
		\STATE Update topic assignments and topic proportions using \eqref{jdtm:lambda:d} and \eqref{jdtm:tau:d}
		\ENDFOR
		\ENDFOR
		\FOR{each shared topic $k$ in $1,2...,K$}
		\FOR{$t$ in $1,2,...,T$}
		\STATE Compute gradients according to \eqref{jdtm:update:topic} and update $\hat{\alpha}_{t,k}$ using conjugate gradient method
		\ENDFOR
		\ENDFOR
		\STATE Update lead-specific and lag-specific topics similarly as shared topics
		\STATE Compute ELBO according to \eqref{jdtm:elbo}
		\ENDWHILE
	\end{algorithmic}
\end{algorithm}

\subsection{An embedding extension}

To handle the problem of large vocabularies, topic models with embeddings have attracted much attentions recently. A notable work is the dynamic embedding topic model \citep{dieng2019the}, which represents dynamic topics in embedding spaces. Specifically, for each topic $k$ at a given time $t=1,...,T$, its word probabilities $\beta_{t,k}$ is represented by an $L$-dimensional vector, i.e., $\beta_{t,k}\in\mathcal{R}^L$. Each word in the vocabulary is also represented by an embedding vector. Suppose the vocabulary has $V$ distinct words, then the word embedding matrix $\rho$ is a $L \times V$ matrix. The probability of each word under each topic is given by the exponentiated inner product between the embedding representations of the word and the topic at the corresponding time, i.e., $p(w_{d,n}=v | z_{d,n}=k, \beta_{t_d,k}) \propto \exp \{\rho_{v}^{\top} \beta_{t_d,k}\}$. Then similarly with DTM, the evolution of topic probabilities is modeled by a Gaussian distribution with variance $\sigma^2$, i.e., $\beta_{t,k} | \beta_{t-1,k}\sim N(\beta_{t-1,k}, \sigma^{2} I)$.

Following the idea of \citep{dieng2019the}, we propose the jointly dynamic embedding topic model (JDETM), which extends JDTM with the embedding techniques. In general, all the notations are the same as JDTM, except for the topic probabilities, which are represented by embeddings. That is, we assume $\alpha_{t,k},\beta_{t,j},\gamma_{t,h} \in \mathcal{R}^L$ for the three types of topics. In JDETM, we marginalize out the topic assignments $\bm{z}$ and regard the Gaussian priors $\eta_t$ and $\kappa_t$ as latent variables. Denote $\bm{\eta}=\{\eta_t:1\leq t \leq T\}$ and $\bm{\kappa}=\{\kappa_t:1\leq t \leq T\}$. Then all latent variables to be estimated in JDETM are $(\bm{\alpha},\bm{\beta},\bm{\gamma},\bm{\theta},\bm{\phi},\bm{\eta},\bm{\kappa})$. Denote $\widetilde{\Lambda}$ be the collection of all hyperparameters. Then the posterior distribution in JDETM is $\widetilde{p}=p(\bm{\alpha},\bm{\beta},\bm{\gamma},\bm{\theta},\bm{\phi},\bm{\eta},\bm{\kappa}|\mathcal{C}_{lead}, \mathcal{C}_{lag},\widetilde{\Lambda})	$.

We still use variational inference for model estimation. Following the inference procedure of DETM \cite{dieng2019the}, we preserve some of the conditional dependencies of the graphical model through the structured variational family \citep{Saul1995Exploiting}. We assume the parameters of the approximated distributions as the functions of the data \citep{Kingma2013Auto}. Specifically, for any document $d$ in the leading corpus with its timestamp $t_d$, we model the variational mean and variance of topic proportions as a function of the latent prior $\eta_{t_d}$ and all words in this document $\textbf{w}_d$. The latent prior $\eta_{t_d}$ is represented by all the previous latent priors $\eta_{1: {t_d}-1}$ and the normalized representations of words in all leading documents $\mathcal{C}_{lead,t}$. The specifications of the variational parameters for the lagged corpus are the same as those for the leading corpus. Denote all the variational parameters as $\widetilde{\Gamma}$. Then, the variational distribution in JDETM has the following form
\begin{equation}
\begin{aligned}
&\widetilde{q}=q(\bm{\alpha},\bm{\beta},\bm{\gamma},\bm{\theta},\bm{\phi},\bm{\eta},\bm{\kappa}| \mathcal{C}_{lead}, \mathcal{C}_{lag},\widetilde{\Gamma})\\
&= \prod_{k=1}^K q(\alpha_{1,k},...,\alpha_{T,k}|\widetilde{\Gamma})\prod_{j=1}^J q(\beta_{1,j},...,\beta_{T,j}|\widetilde{\Gamma})\prod_{h=1}^H q(\gamma_{1,h},...,\gamma_{T,h}|\widetilde{\Gamma})\\
&\times \prod_{t=1}^T\prod_{d=1}^{D_{lead,t}} q(\theta_{d} | \eta_{t}, \mathbf{w}_{d},\widetilde{\Gamma})\times \prod_{t=1}^T q\left(\eta_{t} | \eta_{1: t-1}, \mathcal{C}_{lead,t},\widetilde{\Gamma}\right) \\
&\times \prod_{t=1}^T\prod_{d'=1}^{D_{lag,t}} q(\phi_{d'} | \kappa_{t}, \mathbf{w}_{d'},\widetilde{\Gamma}) \times \prod_{t=1}^T q\left(\kappa_{t} | \kappa_{1: t-1}, \mathcal{C}_{lag,t},\widetilde{\Gamma}\right). \nonumber
\end{aligned}
\end{equation}
Under these assumptions, we model the topic proportions by feed-forward neural networks and the latent priors by LSTM \cite{hochreiter1997long}. Our goal is to maximize the ELBO (i.e., $\text{E}_q\left[\log\widetilde{p}-\log\widetilde{q}\right]$) with respective to the variational parameters $\widetilde{\Gamma}$. With the reparameterization trick in \cite{Kingma2013Auto}, we can obtain gradients by using the back propagation method and then optimize the variational parameters. In particular, we use the method of stochastic gradient descent and set the learning rate by Adam \cite{kingma2015adam}. Algorithm \ref{alg2} summarizes the estimation procedure of JDETM.

\begin{algorithm}[h]
	\caption{Model Estimation Procedure for JDETM}
	\label{alg2}
	\begin{algorithmic}[1]
		\REQUIRE $\mathcal{C}_{lead}$, $\mathcal{C}_{lag}$, $\widetilde{\Lambda}$, $\widetilde{\Gamma}$, $K$, $J$, $H$, and $l$.
		\STATE Initialize all variational parameters
		\WHILE{the ELBO has not converged,}
		\STATE Sample the latent means $\bm{\eta}$ and $\bm{\kappa}$
		\STATE Sample the topic embeddings $\bm{\alpha},\bm{\beta},\bm{\gamma}$
		\STATE Compute the shared topics: $\pi_{t,k}=\text{softmax}(\rho^\top\alpha_{t,k})$
		\STATE Compute the lead-specific topics: $\pi_{t,j}=\text{softmax}(\rho^\top\beta_{t,j})$
		\STATE Compute the lag-specific topics: $\pi_{t,h}=\text{softmax}(\rho^\top\gamma_{t,h})$
		\STATE Obtain a mini-batch of documents
		\FOR{each document $d$ in the mini-batch}
		\IF{$d$ is leading document}
		\STATE Sample the topic proportions $\theta_d \sim q(\theta_d | \eta_{t_d},\mathbf{w}_d)$
		\FOR{each word $n$ in the document}
		\STATE Compute $p(w_{d,n} | \theta_d)=\sum_k\theta_{d_k}\pi_{t_d,k,w_{d,n}}+\sum_j\theta_{d_j}\pi_{t_d,j,w_{d,n}}$
		\ENDFOR
		\ELSE
		\STATE Sample the topic proportions $\phi_{d'} \sim q(\phi_{d'} | \kappa_{t_{d'}},\mathbf{w}_{d'})$
		\FOR{each word $n$ in the document}
		\STATE Compute $p(w_{d_n} | \phi_d)=\sum_k\phi_{d_k}\pi_{t_{d'},k,w_{d_n}}+\sum_h\phi_{d_h}\pi_{t_{d'}-l,h,w_{d'_n}}$
		\ENDFOR
		\ENDIF
		\ENDFOR
		\STATE Estimate the ELBO and its gradient w.r.t. the variational parameters
		\STATE Update the model and variational parameters using Adam
		\ENDWHILE
	\end{algorithmic}
\end{algorithm}

\section{Experiments on Synthetic Data}

\subsection{Experimental design}

To demonstrate the performance of our proposed jointly dynamic topic modeling approach, we design a variety of experiments on synthetic data. Given topics models are all probabilistic generative models, we can generate synthetic documents according to the generative process illustrated in Figure \ref{fig:ourmodel}. Under a fixed time span $T$ and the numbers of three types of topics (i.e., $K$, $J$ and $H$), we first present the generation process for topic proportions, i.e., $\{\theta_d\}$s of the leading corpus and $\{\phi_{d'}\}$s of the lagged corpus. First, we randomly generate the initial values of the topic proportion means $\eta_0$ and $\kappa_0$ from the standard normal distribution $N(0,1)$. Second, for time $1 \leq t \leq T$, the topic proportion means $\eta_t$ and $\kappa_t$ are generated from normal distributions $N(\eta_{t-1},1)$ and $N(\kappa_{t-1},1)$, respectively. Last, given all topic proportion means, the topic proportions $\{\theta_d\}$s and $\{\phi_{d'}\}$s are generated from log-normal distributions $LN(\eta_{t_d},1)$ and $LN(\kappa_{t_{d'}},1)$, respectively. Next, we generate documents under two scenarios to investigate the performance of JDTM and JDETM.

{\sc Scenario 1.} In this scenario, we generate documents under the generation procedure of JDTM.
Specifically, we set the time span $T=10$ and the vocabulary size $V=1000$. For $K$ shared topics, $J$ lead-specific topics and $H$ lag-specific topics, the topic probabilities are initially generated from the standard normal distribution $N(0,1)$ and then evolve with Gaussian noises, e.g., $\alpha_{t,k} \sim N(\alpha_{t-1,k},1)$. For time $1 \leq t \leq T$, the number of documents $D_{lead,t}$ and $D_{lag,t}$ are both generated from the Poisson distribution $Po(100)+50$. For each document in the leading corpus or the lagged corpus, the number of words is generated from the Poisson distribution $Po(50)+50$.

{\sc Scenario 2.} In this scenario, we generate large document corpus under the generation procedure of JDETM. To obtain the topic embeddings and word embeddings, we follow \citep{dieng2019the} to apply DETM on the ACL dataset with a total of 50 topics. Then, we split the 50 topics into $J$ lead-specific topics, $H$ lag-specific topics, and $K$ shared topics. The JS-divergence is calculated between distributions of lead-specific topics and lag-specific topics to guarantee certain differences between the two types of topics. Then, following the ACL dataset, we set the time span $T=31$ and the vocabulary size $V=17,794$. For time $1 \leq t \leq T$, the number of documents $D_{lead,t}$ and $D_{lag,t}$ are both generated from the Poisson distribution $Po(400)+100$. For each document in the leading corpus or the lagged corpus, the number of words is generated from the Poisson distribution $Po(300)+50$.

In the proposed model, the lead-lag relationship is characterized by the \emph{shared topics}. Therefore, the proportion of shared topics (i.e., $K$) among all topics (i.e., $K+J+H$) can represent the strength of lead-lag relationship. To consider different strengths of lead-lag relationship, we change the numbers of three types of topics to mimic the situation of \emph{no} lead-lag relationship, \emph{weak} lead-lag relationship and \emph{strong} lead-lag relationship. In each situation, we consider the lagged time period $l=1,3,5$. As a result, we have a combination of 2 scenarios $\times$ 3 situations $\times$ 3 lagged values $=18$ experimental settings. In each setting, we repeat the experiment for 20 times.

\subsection{Evaluation metrics and comparison methods}
We compare our model with two approaches. In the first approach, we apply the DTM or the DETM to fit the leading corpus and lagged corpus, separately. We denote this separately modeling approach as $\text{DTM}_{\text{s}}$ and $\text{DETM}_{\text{s}}$. In this approach, the numbers of topics set for the leading corpus and the lagged corpus are $K+J$ and $K+H$, respectively. In the second approach, we combine the leading corpus and the lagged corpus together and then apply the DTM or DETM. We denote this modeling approach as $\text{DTM}_{\text{c}}$ and $\text{DETM}_{\text{c}}$. In this approach, the number of topics set for the combined corpora is $K+J+H$.

We use the document completion perplexity \citep{Wallach2009Eval} to evaluate the model performance.
The task of document completion aims to estimate the probability of the second half of a document given its first half.
Let $d^{(1)}$ be the first half of document $d$ and $d^{(2)}$ be the second half. Then the completion perplexity of corpus $\mathcal{C}$ is computed as follows,
\begin{equation}
	PPL(\mathcal{C}) = \exp\Big(-\frac{\sum_{d} \log p(d^{(2)}|d^{(1)},\mathcal{M})}{\sum_{d} N_d}\Big),
\end{equation}
where $\mathcal{M}$ is the fitted model and $p(d^{(2)}|d^{(1)},\mathcal{M}) = \prod^{N_d}_{i=1}p(z_i|d^{(1)},\mathcal{M})p(w_i|z_i)$. The lower the perplexity, the better the model performance.

\subsection{Results}

Table \ref{tab:results_s1} shows the averaged document completion perplexities among 20 experiments in each experimental setting in scenario 1. For each method, we compare the perplexity on the leading corpus, the lagged corpus, as well as the combination of two corpora. For easy understanding, we also calculate the lift proportions between the perplexities in JDTM and its competitors. As shown by Table \ref{tab:results_s1}, under the situation of \emph{no} lead-lag relationship, the JDTM obtains comparable results with DTMs. In addition, both the two methods have largely outperformed DTMc. However, when there exists \emph{weak} lead-lag relationship or \emph{strong} lead-lag relationship, the JDTM can outperform DTMs and DTMc. Particularly, the JDTM has shown greater advantages in the lagged corpus than in the leading corpus. In addition, with the stronger lead-lag relationship, the advantages of JDTM on the lagged corpus becomes more obvious.

\begin{table}[h]
	\caption{Quantitative performance of scenario 1 as measured by perplexity.}
    \footnotesize
	\label{tab:results_s1}
    \renewcommand\arraystretch{1.2}
	\begin{center}
		\begin{threeparttable}[b]
			\begin{tabular}{cccccccc}
				\hline
				Settings &Lag	&Corpus	&DTMs	&DTMc	&JDTM	&V.S. DTMs	&V.S. DTMc\\
                \hline
                \multirow{9}{*}{\tabincell{c}{\emph{No} lead-lag \\ $K=0$\\ $J=5$\\ $H=5$}}
                &\multirow{3}{*}{1}	&Lead	&426.21	&626.74	&426.09	&0.03\%	&32.01\%	\\
                    &	&Lag	&442.69	&583.38	&447.28	&-1.04\%	&23.33\%	\\
                    &	&Combine	&434.45	&605.06	&436.69	&-0.51\%	&27.83\%	\\
                    \cline{2-8}
                    &\multirow{3}{*}{3}	&Lead	&499.34	&727.59	&500.08	&-0.15\%	&31.27\%	\\
                    &	&Lag	&357.21	&485.31	&364.05	&-1.91\%	&24.99\%	\\
                    &	&Combine	&428.28	&606.45	&432.07	&-0.88\%	&28.76\%	\\
                    \cline{2-8}
                    &\multirow{3}{*}{5}	&Lead	&444.95	&640.07	&444.48	&0.11\%	&30.56\%	\\
                    &	&Lag	&243.25	&337.41	&246.52	&-1.34\%	&26.94\%	\\
                    &	&Combine	&344.10	&488.74	&345.50	&-0.41\%	&29.31\%	\\
                \hline
                \multirow{9}{*}{\tabincell{c}{\emph{Weak} lead-lag \\ $K=5$\\ $J=5$\\ $H=5$}}
                    &\multirow{3}{*}{1}	&Lead	&506.93	&607.87	&504.45	&0.49\%	&17.01\%	\\
                    &	&Lag	&446.05	&505.50	&404.17	&9.39\%	&20.04\%	\\
                    &	&Combine	&476.49	&556.68	&454.31	&4.66\%	&18.39\%	\\
                    \cline{2-8}
                    &\multirow{3}{*}{3}	&Lead	&415.57	&498.82	&411.97	&0.87\%	&17.41\%	\\
                    &	&Lag	&547.60	&622.38	&459.81	&16.03\%	&26.12\%	\\
                    &	&Combine	&481.59	&560.60	&435.89	&9.49\%	&22.25\%	\\
                    \cline{2-8}
                    &\multirow{3}{*}{5}	&Lead	&413.94	&508.98	&412.26	&0.41\%	&19.00\%	\\
                    &	&Lag	&501.27	&574.94	&385.98	&23.00\%	&32.87\%	\\
                    &	&Combine	&457.61	&541.96	&399.12	&12.78\%	&26.36\%	\\
                \hline
                \multirow{9}{*}{\tabincell{c}{\emph{Strong} lead-lag \\ $K=9$\\ $J=1$\\ $H=1$}}
                    &\multirow{3}{*}{1}	&Lead	&452.84	&598.62	&451.88	&0.21\%	&24.51\%	\\
                    &	&Lag	&332.44	&399.14	&302.21	&9.09\%	&24.28\%	\\
                    &	&Combine	&392.64	&498.88	&377.05	&3.97\%	&24.42\%	\\
                    \cline{2-8}
                    &\multirow{3}{*}{3}	&Lead	&450.40	&564.07	&450.16	&0.05\%	&20.19\%	\\
                    &	&Lag	&421.50	&487.22	&303.55	&27.98\%	&37.70\%	\\
                    &	&Combine	&435.95	&525.65	&376.86	&13.55\%	&28.31\%	\\
                    \cline{2-8}
                    &\multirow{3}{*}{5}	&Lead	&533.91	&668.44	&531.18	&0.51\%	&20.54\%	\\
                    &	&Lag	&511.62	&584.36	&360.32	&29.57\%	&38.34\%	\\
                    &	&Combine	&522.77	&626.40	&445.75	&14.73\%	&28.84\%	\\

			\hline
			\end{tabular}
		\end{threeparttable}
	\end{center}
\end{table}

Table \ref{tab:results_s2} shows the averaged document completion perplexities in scenario 2. In general, the results in Table \ref{tab:results_s2} show similar trends with those in Table \ref{tab:results_s1}. When there is \emph{no} lead-lag relationship, JDETM and DETMs have outperformed DETMc. When there exists \emph{weak} lead-lag relationship or \emph{strong} lead-lag relationship, the JDETM shows advantages over the other methods. Specifically, regarding the estimation performance in the leading corpus, the JDETM can achieve the best or comparable performance with its competitors. In terms of the estimation performance in the lagged corpus or the combination of the two corpora, the JDETM can always outperform all the other models, and its advantages are more obvious.
\begin{table}[h]
	\caption{Quantitative performance of scenario 2 as measured by perplexity.}
    \footnotesize
	\label{tab:results_s2}
    \renewcommand\arraystretch{1.2}
	\begin{center}
		\begin{threeparttable}[b]
			\begin{tabular}{cccccccc}
				\hline
				Settings &Lag	&Corpus	&DETMs	&DETMc	&JDETM	&V.S. DETMs	&V.S. DETMc\\
                \hline
                \multirow{9}{*}{\tabincell{c}{\emph{No} lead-lag \\ $K=0$\\ $J=10$\\ $H=10$}}
                &\multirow{3}{*}{1}	&Lead	&656.50	&717.19	&665.33	&-1.35\%	&7.23\%	\\
                    &	&Lag	&650.96	&698.08	&661.63	&-1.64\%	&5.22\%	\\
                    &	&Combine	&653.73	&707.64	&663.48	&-1.49\%	&6.24\%	\\
                    \cline{2-8}
                    &\multirow{3}{*}{3}	&Lead	&739.99	&793.32	&749.13	&-1.24\%	&5.57\%	\\
                    &	&Lag	&652.05	&687.49	&659.22	&-1.10\%	&4.11\%	\\
                    &	&Combine	&696.02	&740.41	&704.18	&-1.17\%	&4.89\%	\\
                    \cline{2-8}
                    &\multirow{3}{*}{5}	&Lead	&767.81	&826.90	&784.47	&-2.17\%	&5.13\%	\\
                    &	&Lag	&702.19	&752.67	&711.99	&-1.40\%	&5.40\%	\\
                    &	&Combine	&735.00	&789.79	&748.23	&-1.80\%	&5.26\%	\\
                \hline
                \multirow{9}{*}{\tabincell{c}{\emph{Weak} lead-lag \\ $K=5$\\ $J=20$\\ $H=20$}}
                &\multirow{3}{*}{1}	&Lead	&867.29	&912.11	&880.59	&-1.53\%	&3.46\%	\\
                    &	&Lag	&855.35	&907.42	&841.08	&1.67\%	&7.31\%	\\
                    &	&Combine	&861.32	&909.77	&860.84	&0.06\%	&5.38\%	\\
                \cline{2-8}
                    &\multirow{3}{*}{3}	&Lead	&996.15	&1060.25	&1012.39	&-1.63\%	&4.51\%	\\
                    &	&Lag	&1130.19	&1208.71	&1067.49	&5.55\%	&11.68\%	\\
                    &	&Combine	&1063.17	&1134.48	&1039.94	&2.18\%	&8.33\%	\\
                \cline{2-8}
                    &\multirow{3}{*}{5}	&Lead	&874.87	&913.71	&883.60	&-1.00\%	&3.30\%	\\
                    &	&Lag	&914.75	&966.85	&813.24	&11.10\%	&15.89\%	\\
                    &	&Combine	&894.81	&940.28	&848.42	&5.18\%	&9.77\%	\\
                \hline
                \multirow{9}{*}{\tabincell{c}{\emph{Strong} lead-lag \\ $K=35$\\ $J=5$\\ $H=5$}}
                &\multirow{3}{*}{1}	&Lead	&877.66	&923.44	&896.44	&-2.14\%	&2.92\%	\\
                    &	&Lag	&866.42	&898.18	&831.70	&4.01\%	&7.40\%	\\
                    &	&Combine	&872.04	&910.81	&864.07	&0.91\%	&5.13\%	\\
                \cline{2-8}
                    &\multirow{3}{*}{3}	&Lead	&982.41	&1041.97	&1010.84	&-2.89\%	&2.99\%	\\
                    &	&Lag	&952.79	&1004.33	&865.39	&9.17\%	&13.83\%	\\
                    &	&Combine	&967.60	&1023.15	&938.12	&3.05\%	&8.31\%	\\
                \cline{2-8}
                    &\multirow{3}{*}{5}	&Lead	&935.87	&978.81	&949.67	&-1.47\%	&2.98\%	\\
                    &	&Lag	&769.97	&810.20	&638.11	&17.12\%	&21.24\%	\\
                    &	&Combine	&852.92	&894.51	&793.89	&6.92\%	&11.25\%	\\
			\hline
			\end{tabular}
		\end{threeparttable}
	\end{center}
\end{table}

\section{Experiments on Real Data}

\subsection{Data description}
In this section, we try to apply the jointly dynamic topic model to investigate the influences of academic papers on students' scientific researches. To this end, we take the field of \emph{statistics} as an example, and collect two text corpora as follows. The first text corpus is statistical papers published on top journals. Specifically, according to the number of journal citations in Web of Science (i.e., the Journal Citation Report in 2019), ten journals with the highest number of citations are selected under the category of \emph{Statistics and Probability}. Then, all papers published on these ten journals from 2005 to 2019 are crawled. It results in a total of 12,348 papers. For each paper, its authors, published time, abstract and keywords are collected. The abstracts are then used as the leading corpus, which is denoted as EN. The second text corpus is the Chinese graduation theses majored in statistics. Specifically, we focus on the Chinese master graduation theses majored in \emph{Statistics or Probability Theory} and \emph{Mathematical Statistics}. We crawl these theses published between 2005 to 2019 from the CNKI Database.\footnote{https://epub.cnki.net/kns/brief/result.aspx?dbPrefix=CDMD} It results in a total of 12,903 theses. For each thesis, its author, published time, graduation school, abstract and keywords are collected. The abstracts in all graduation theses are considered as the lagged corpus, which is denoted as CN. The ten selected journals for the leading corpus are listed in Appendix A. For illustration purpose, Appendix A also presents the names of ten universities with the largest number of theses in the lagged corpus.

For each corpus, we first preprocess the text documents by removing stop words. Given the lagged corpus is written in Chinese, word segmentation using an open-source package \emph{Jieba} is performed. Then, we remove words with document frequencies higher than 70\% or appearing in less than 10 documents from both corpora. Finally, to match the written language in the leading corpus and lagged corpus, we apply the Baidu Translator\footnote{https://fanyi.baidu.com/} to translate Chinese words into English. After all these preprocessing steps, the total vocabulary size is $V=8,611$.

\begin{figure}[h]
	\centering
	\includegraphics[width=0.6\linewidth]{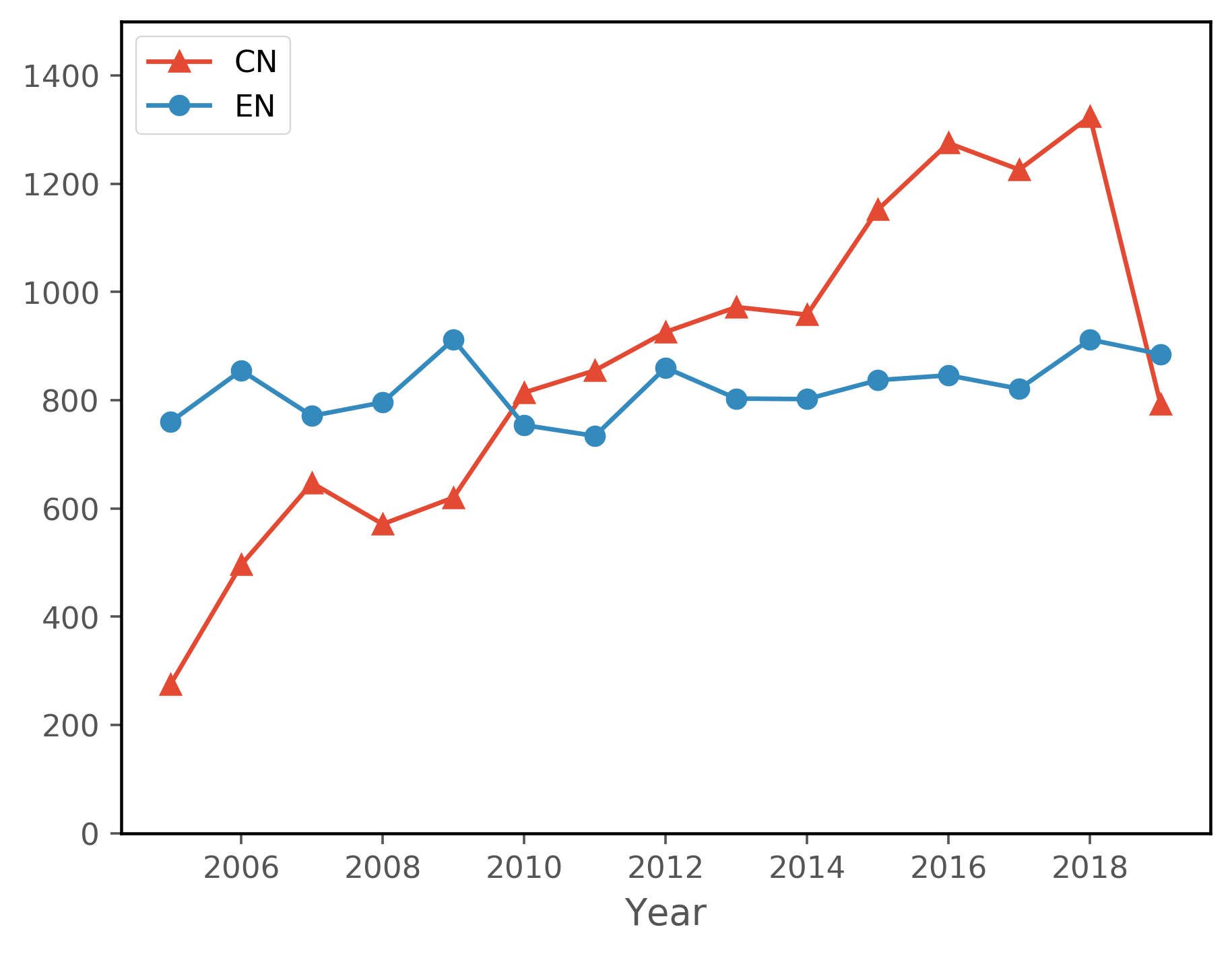}
	\caption{The number of documents in the leading corpus (EN) and lagged corpus (CN) from 2005 to 2019. The EN corpus has a stable trend in the number of documents, while the CN corpus has an obvious increasing trend in the number of documents.}
	\label{fig:doc_nums}
\end{figure}

We organize the two corpora in the yearly format, which results in 15 years in total. The numbers of documents in the two corpora in each year are shown in Figure \ref{fig:doc_nums}. As shown, the number of published papers is relatively stable across year, which is mainly due to the journals' issue policies. On the other hand, there is an obvious increasing trend in the number of graduation theses. This indicates the enrollment of gradate students is expanding in China.

\begin{figure}[htbp]
	\centering
	\includegraphics[width=0.98\linewidth]{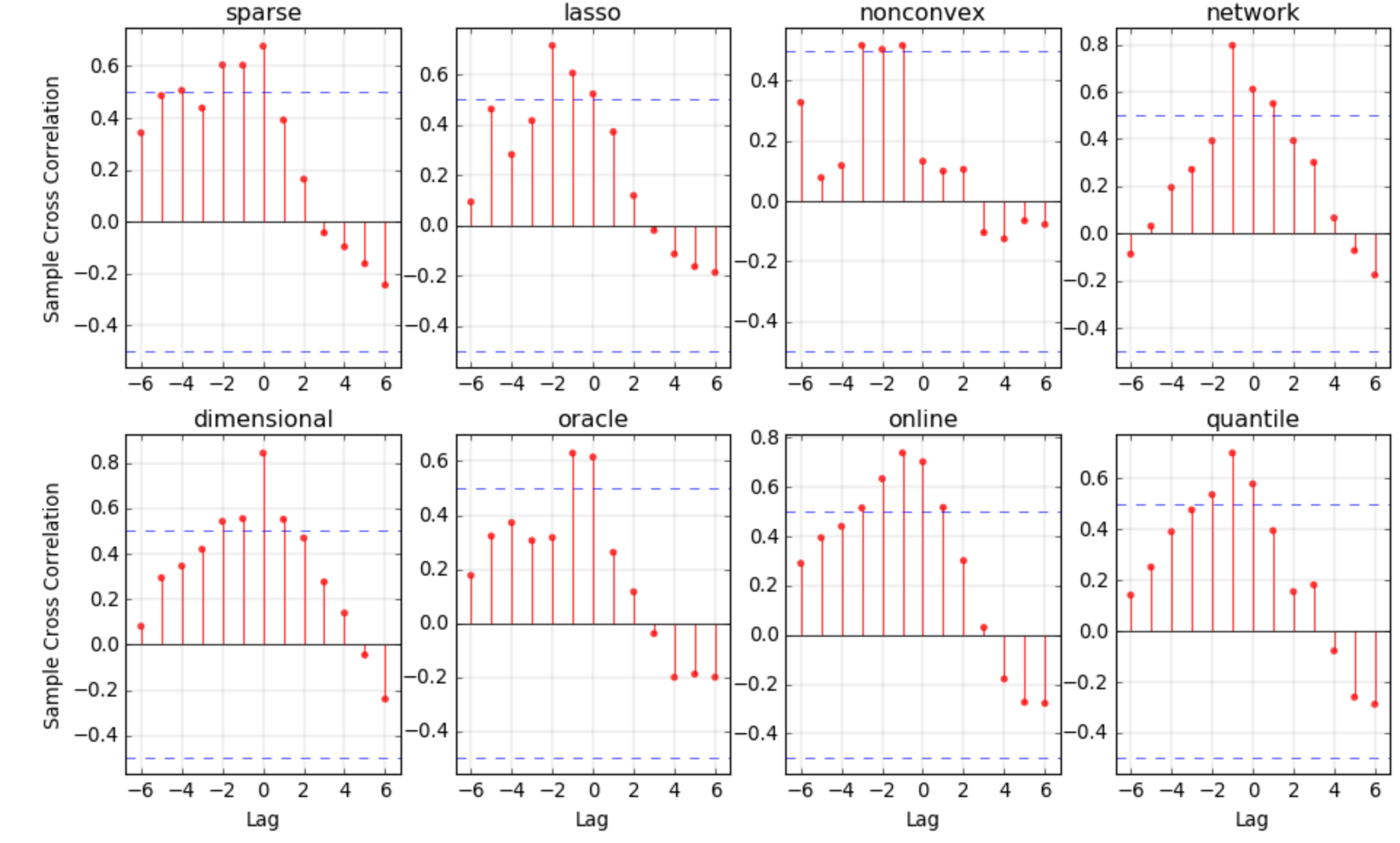}
	\caption{The lagged cross correlation plots for the yearly frequencies of eight words in the journal corpus EN and the graduation theses corpus CN. The bar is the correlation coefficient at each lag, while the two horizontal dotted lines represent the rough confidence interval with 95\% confidence level. The significant correlation coefficients can be found for each word around lag -2 to 2.}
	\label{fig:ccf_plot}
\end{figure}

We then explore the potential lead-lag relationship in the two corpora by applying some time series tools. We first choose eight example words, which represent classic topics in statistics. They are, ``sparse", ``lasso", ``nonconvex", ``network", ``dimensional", ``oracle", ``online", and ``quantile", respectively. For each word $w$, we calculate its yearly frequencies in the two corpora, and regard them as two time series which are denoted by $\text{EN}_{w}$ and $\text{CN}_{w}$. We then draw the lagged cross correlation plot \citep{2008Time} for the two time series. The corresponding results are shown in Figure \ref{fig:ccf_plot}. It is obvious that, all the eight words have significant lagged correlation coefficients around lag -2 to 2, as their coefficient bars exceed the corresponding confidence intervals. These results suggest that, the time series of the EN corpus predates uptrend that of the CN corpus. These results motivate us to further explore the lead-lag relationship between the two corpora by the jointly dynamic topic model.

%In addition, we conduct the Granger causality test for the time series of word frequencies in the two corpora, whose results are shown in Table XXX. As shown,....

\subsection{Experimental results using JDETM}

We then apply the jointly dynamic topic modeling approach to explore the relationship between the statistical papers and graduation theses. Given the two text corpora are of large scale, we choose the JDETM for analysis. For illustration purpose, we set $K=10$ shared topics, $J=10$ lead-specific topics, and $H=10$ lag-specific topics. Following \cite{dieng2019the}, we set the hyperparameters as $\sigma^{2}_{k} = \sigma^{2}_{j} = \sigma^{2}_{h} = 0.005$ and $\varpi_f^2=\varpi_l^2=1$. To evaluate the performance of JDETM, we also take $\text{JDETM}_{\text{s}}$ and $\text{JDETM}_{\text{c}}$ as competitors. The number of topics set in $\text{JDETM}_{\text{s}}$ and $\text{JDETM}_{\text{c}}$ are 20 and 30, respectively. To explore the lead-lag relationship in the two corpora, we consider the lagged time period as $l=1,2,3$.

We split the leading corpus and lagged corpus into three parts, i.e., the training dataset (85\%), the validation dataset (5\%) and the test dataset (10\%). The training dataset and validation dataset are used for model estimation, while the test dataset is used for model evaluation. For all embedding topic models, we use randomly initialized 300-dimensional word embeddings. We then employ stochastic variational inference with a batch size of 200 documents and take a training procedure for 400 epochs. For the topic proportions $\{\theta_d\}$s and $\{\phi_{d'}\}$s, we use a fully connected feed-forward network with 2 layers of 800 hidden units and ReLU activations.
Given the length of documents is not long, we fix the means of latent proportions as $\eta=\kappa=0.1$, which can reduce the model complexity and prevent overfitting. The learning rates in all methods are fixed as 0.001.

Table \ref{tab:results_JDETM} shows the performance of JDETM and its competitors on the test dataset. We can draw the following conclusions. First, the method of $\text{DETM}_{\text{c}}$ achieves the worse performance in both the leading corpus and lagged corpus. This finding suggests that simply combining the two types of documents would not improve the performance of topic learning, even though more information is considered. Second, the JDETM has achieved the best performance in the lagged corpus, while kept comparable performance with $\text{DETM}_{\text{s}}$ in the leading corpus. These results verify the superiority of JDETM in modeling the lagged corpus. It is also notable that, compared with $\text{DETM}_{\text{s}}$, JDETM only needs to be trained once, which is more convenient for practical applications.
\begin{table}[h]
	\caption{Quantitative performance on the real dataset as measured by perplexity. ``*'' and ``**'' denote the top two method with best performance in each corpus.}
	\label{tab:results_JDETM}
	\begin{center}
		\begin{threeparttable}[b]
			\begin{tabular}{ccccc}
				\hline
				Corpus & Lag &$\text{DETM}_{\text{s}}$ &$\text{DETM}_{\text{c}}$ & $\text{JDETM}$ \\
				\hline
				\multirow{3}{*}{Lead} &1 &\multirow{3}{*}{1356.4*} &\multirow{3}{*}{1642.2} &1382.7**\\
                &2 & & &1407.7\\
                &3 & & &1418.9\\
                \hline
				\multirow{3}{*}{Lag} &1 &\multirow{3}{*}{856.3} &\multirow{3}{*}{833.5} &824.6**\\
                &2 & & &810.8*\\
                &3 & & &839.5\\
                \hline
			\end{tabular}
		\end{threeparttable}
	\end{center}
\end{table}

\subsection{Specific and shared patterns in two corpora}

In this subsection, we try to investigate the specific patterns and shared patterns in the two corpora. As we mentioned before, the shared topics capture the similarities of the two corpora, while the lead-specific topics and lag-specific topics capture the characteristics in each single corpus. Therefore, by analyzing the top words with highest probabilities in the three types of topics, we can have a general idea about the specific and shared patterns of the two corpora. To illustrate this idea, we randomly choose two lead-specific topics (``Fuzzy System" and ``High-dimensional Statistics"), two lag-specific topics (``Inequality Problems in China" and ``Evaluation Index"), and four shared topics (``Economic Development", ``Parameter Estimation", ``Financial Markets", and ``Model Selection") under the JDETM with $l=1$ as examples. We name the topics by summarizing the meanings of words with high probabilities under each topic. Then, we select the top six words with the highest probabilities in each example topic, and show their probability trends across time in Figure

As shown, the top words in the chosen topics are in agreement with the characterisers of the two corpora. For example, documents related to fuzzy system mainly appear in the journal \textit{Fuzzy Sets and Systems}. Therefore, the topic ``Fuzzy System" has been recognized as a lead-specific topic. In addition, the topics ``Inequality Problems in China" and ``Evolution Index" are mainly discussed by Chinese students in their graduation theses, and thus these two topics appear as lag-specific topics. Regarding the four shared topics, they are common themes discussed in the field of statistics, such as the parameter estimation and financial markets. As a consequence, these shared topics recognized by JDETM reflect the common interests of the two corpora.
\begin{figure}[h]
	\centering
    \footnotesize
	\includegraphics[width=1\textwidth]{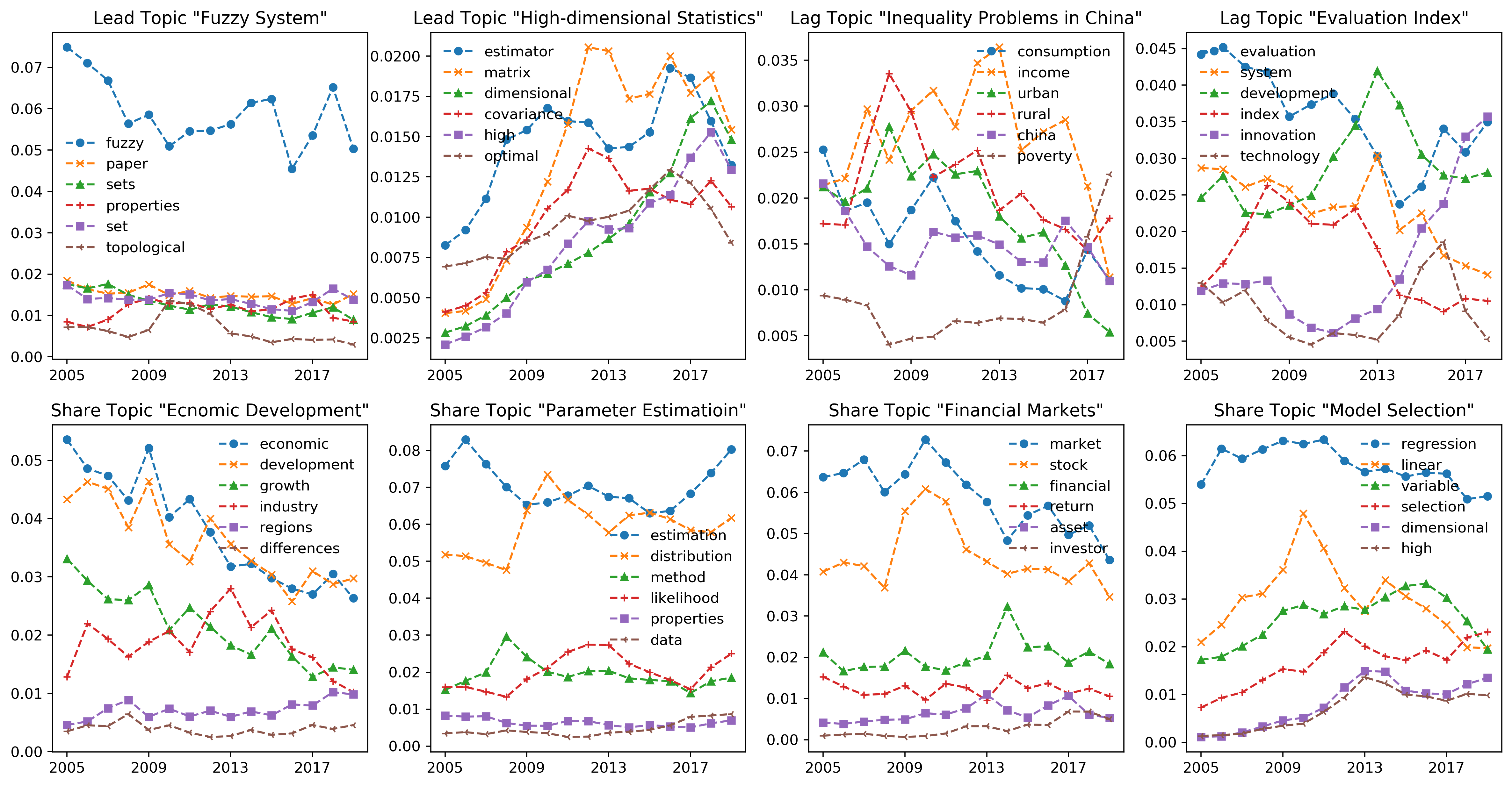}
	\caption{The time trend of top six words with highest probabilities in two lead-specific topics, two lag-specific topics, and four shared topics. In general, the top words in the lead-specific topics can reflect the characteristics of the leading corpus, while the lag-specific topics can reflect the characteristics of the lagged corpus. As for the shared topics, their top words reflect the common themes discussed by both corpora.}
	\label{fig:word_evol_JDETM}
\end{figure}

Next, we take the eight topics shown in Figure \ref{fig:word_evol_JDETM} as examples to further demonstrate the lead-lag relationship recognized by JDETM. Note that we have found the top six words with the highest probabilities for each topic, which represent the main meanings of the topics. We then test whether the causal relationship exists for each word in the two corpora. Specifically, for each word $w$, we regard its yearly frequencies $\text{EN}_{w}$ and $\text{CN}_{w}$ as two time series. Then we test the causality of the two time series via the convergent cross mapping (CCM) method \cite{sugihara2012detecting}. The main idea of the CCM method is that, if the time series $\text{EN}_{w}$ causally influences the time series $\text{CN}_{w}$, the historical records of $\text{EN}_{w}$ can be reliably predicted through the manifold generated from $\text{CN}_{w}$. Thus we can test for causation through the correlation between the true values and predicted values of $\text{EN}_{w}$. We apply the CCM method for each word in the eight topics, and the detailed results are shown in Table \ref{tab:words_CCM}. As shown, most top words under the four shared topics (14 out of 24) have significantly causal relationships between the two corpora. On the contrary, most top words under the lead-specific topics and lag-specific topics are insignificant (only 4 out of 24 are significant) or even unpredictable (denoted by ``-''). These results imply the existence of the underlying causality in the two corpora, which are consistent with the lead-lag relationship discovered by JDETM.

\begin{table}[h]
	\caption{The results of CCM method for each top word in the eight selected topics. For each word, the CCM statistics and the test $p$-value are reported. We also use ``*'' to denote the significant causal relationship at the confidence level 95\%. The negative CCM values imply no prediction skill and thus are denoted by ``-''.}
\footnotesize
	\label{tab:words_CCM}
	\begin{tabular}{lccclcc}
    \hline
    \textbf{Top Words} &\textbf{CCM} &\textbf{$p$-value} & &\textbf{Top Words} &\textbf{CCM} &\textbf{$p$-value}\\
    \cline{1-3}
    \cline{5-7}
    \multicolumn{3}{c}{Lead: \emph{Fuzzy System}} & & \multicolumn{3}{c}{Lead: \emph{High-dimensional Statistics}}\\
    \cline{1-3}
    \cline{5-7}
    fuzzy           & 0.279        & 0.356            && estimator       & -            & -                \\
	paper           & 0.399        & 0.224            && matrix          & -            & -                \\
	sets            & 0.315        & 0.273            && dimensional     & 0.848        & 0.001*           \\
	properties      & 0.567        & 0.043*           && covariance      & -            & -                \\
	set             & -            & -                && high            & 0.820         & 0.001*           \\
	topological     & -            & -                && optimal         & 0.356        & 0.233            \\
    \hline
    \multicolumn{3}{c}{Lag: \emph{Inequality Problems in China}} & & \multicolumn{3}{c}{Lag: \emph{Evaluation Index}}\\
    \cline{1-3}
    \cline{5-7}
    consumption     & -            & -                && evaluation      & 0.126        & 0.682            \\
	income          & -            & -                && system          & -            & -                \\
	urban           & 0.390         & 0.168            && development     & 0.540         & 0.046*           \\
	rural           & 0.118        & 0.730             && index           & -            & -                \\
	china           & 0.410         & 0.146            && innovation      & 0.292        & 0.310             \\
	poverty         & 0.079        & 0.788            && technology      & -            & -                \\
    \hline
    \multicolumn{3}{c}{Share: \emph{Economic Development}} & & \multicolumn{3}{c}{Share: \emph{Parameter Estimation}}\\
    \cline{1-3}
    \cline{5-7}
    economic        & 0.535        & 0.090             && estimation      & 0.243        & 0.403            \\
	development     & 0.540         & 0.046*           && distribution    & 0.118        & 0.687            \\
	growth          & -            & -                && method          & 0.690         & 0.019*           \\
	industry        & 0.565        & 0.044*           && properties      & 0.567        & 0.043*           \\
	differences     & 0.585        & 0.028*           && likelihood      & 0.725        & 0.003*           \\
	regions         & 0.587        & 0.035*           && data            & 0.616        & 0.025*           \\
    \hline
    \multicolumn{3}{c}{Share: \emph{Financial Markets}} & & \multicolumn{3}{c}{Share: \emph{Model Selection}}\\
    \cline{1-3}
    \cline{5-7}
    market          & 0.518        & 0.069            && regression      & 0.043        & 0.885            \\
	stock           & 0.579        & 0.030*            && linear          & -            & -                \\
	financial       & -            & -                && variable        & 0.149        & 0.611            \\
	return          & 0.651        & 0.030*            && selection       & 0.289        & 0.317            \\
	asset           & 0.687        & 0.010*            && dimensional     & 0.848        & 0.001*           \\
	investor        & 0.584        & 0.028*           && high            & 0.820        & 0.001*           \\
	\hline
	\end{tabular}
\end{table}

\subsection{The recognized lead-lag relationship}

\begin{figure}[h]
	\centering
	\subfigure[Lag=1]{
		\includegraphics[width=0.45\textwidth]{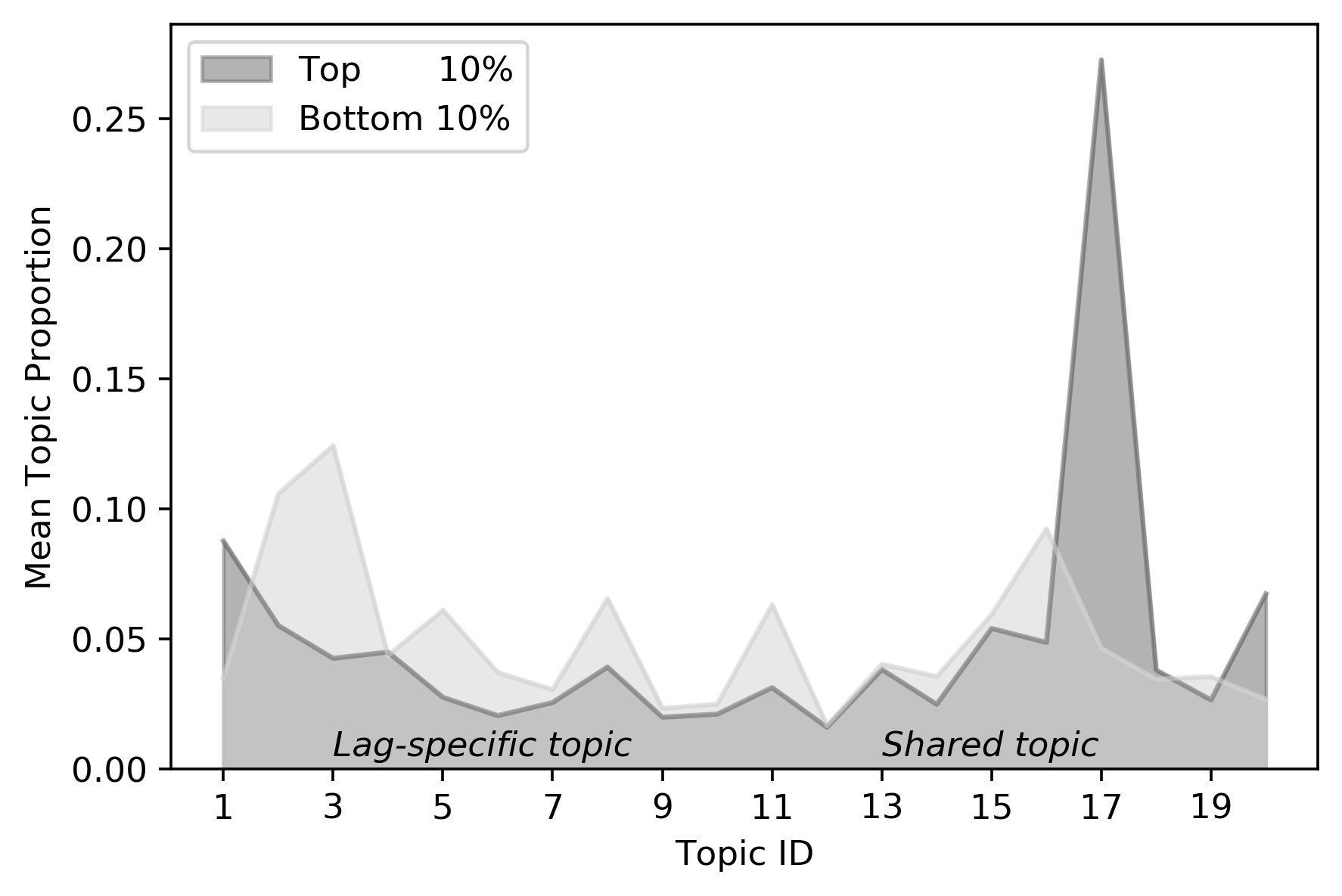}
	}
	\subfigure[Lag=2]{
		\includegraphics[width=0.45\textwidth]{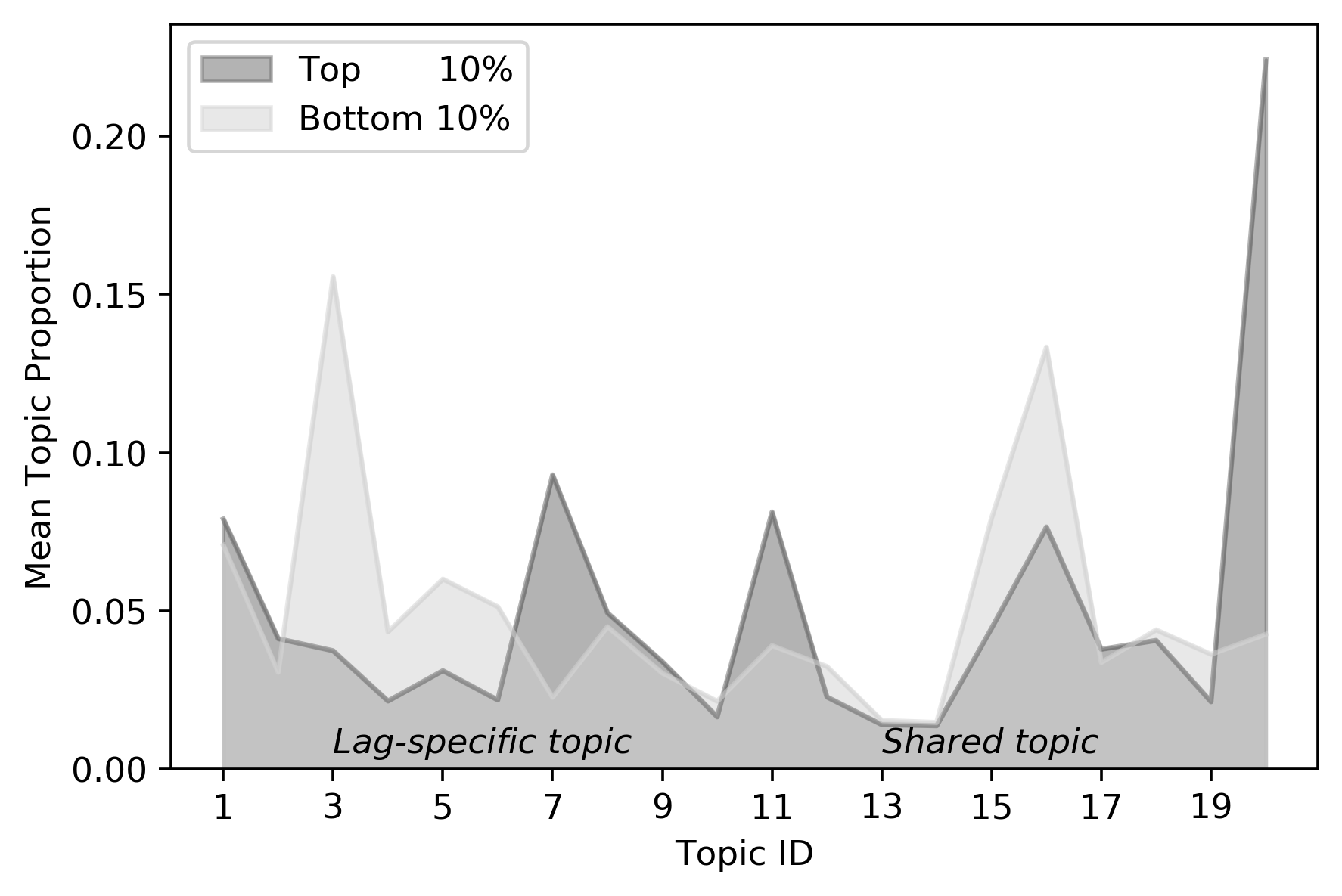}
	}
	\caption{The averaged topic proportions of top 10\% test documents or bottom 10\% test documents in the lagged corpus. In the x-axis, topics 1 to 10 refer to the lag-specific topics while topics 11 to 20 refer to the shared topics. As shown, the top 10\% documents usually have higher probabilities on the shared topics; while the bottom 10\% documents have relatively lower probabilities on the shared topics. }
	\label{fig:mean_topic_prop}
\end{figure}

We focus on the recognized lead-lag relationship between the two corpora in this subsection. First, we try to verify the idea that, the existence of lead-lag relationship can improve the performance of topic modeling in the lagged corpus. Recall that we have applied the JDETM and $\text{DETM}_{\text{s}}$ on the academic corpora, and calculated the perplexity for each test document under the two models. Then, we compute the perplexity difference between the two methods for each test document in the lagged corpus. Next, we select test documents, whose perplexity differences are among the top 10\% or bottom 10\% in the lagged corpus. In other words, the JDETM has performed much better than $\text{DETM}_{\text{s}}$ on the top 10\% documents; while achieved much worse performance than $\text{DETM}_{\text{s}}$ on the bottom 10\% documents. We take JDETM with the lagged time period equal to 1 and 2 for illustration. Figure \ref{fig:mean_topic_prop} presents the averaged topic proportions of these selected documents. Note that all these selected documents are in the lagged corpus, therefore they only have topic proportions on the lag-specific topics as well as the shared topics. As shown, the top 10\% documents usually have higher probabilities on the shared topics; while the bottom 10\% documents have relatively lower probabilities on the shared topics. Therefore, the shared topics contribute to the topic representations of documents (as measured by perplexity) in JDETM. Further recall the lead-lag relationship between the leading corpus and lagged corpus is characterized by the shared topics. These results demonstrate that, by recognizing the lead-lag relationship embedded in the shared topics, the JDETM can improve the quality of topic representations in the lagged corpus.

\begin{figure}[h]
	\centering
	\subfigure[Shared Topic ``Stochastic Process'']{
		\includegraphics[width=0.9\textwidth]{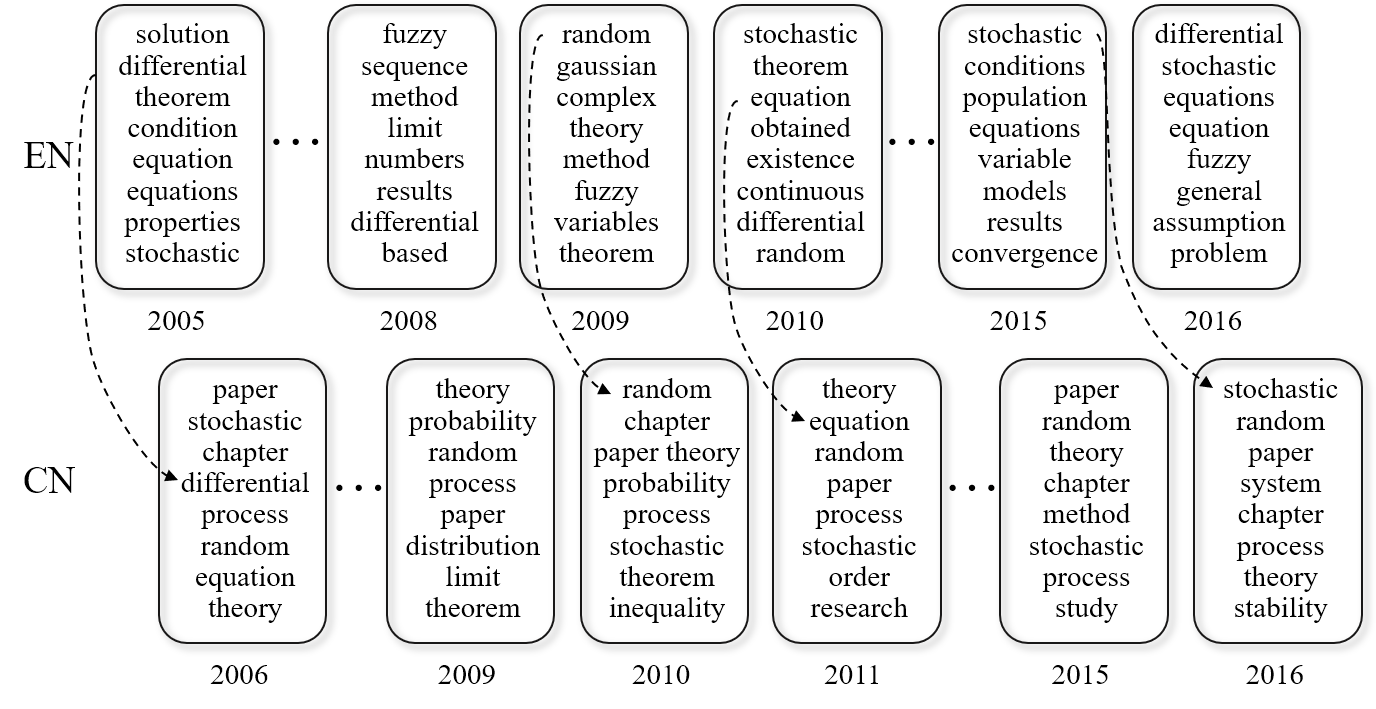}
	}
	\subfigure[Shared Topic ``Model Selection'']{
		\includegraphics[width=0.9\textwidth]{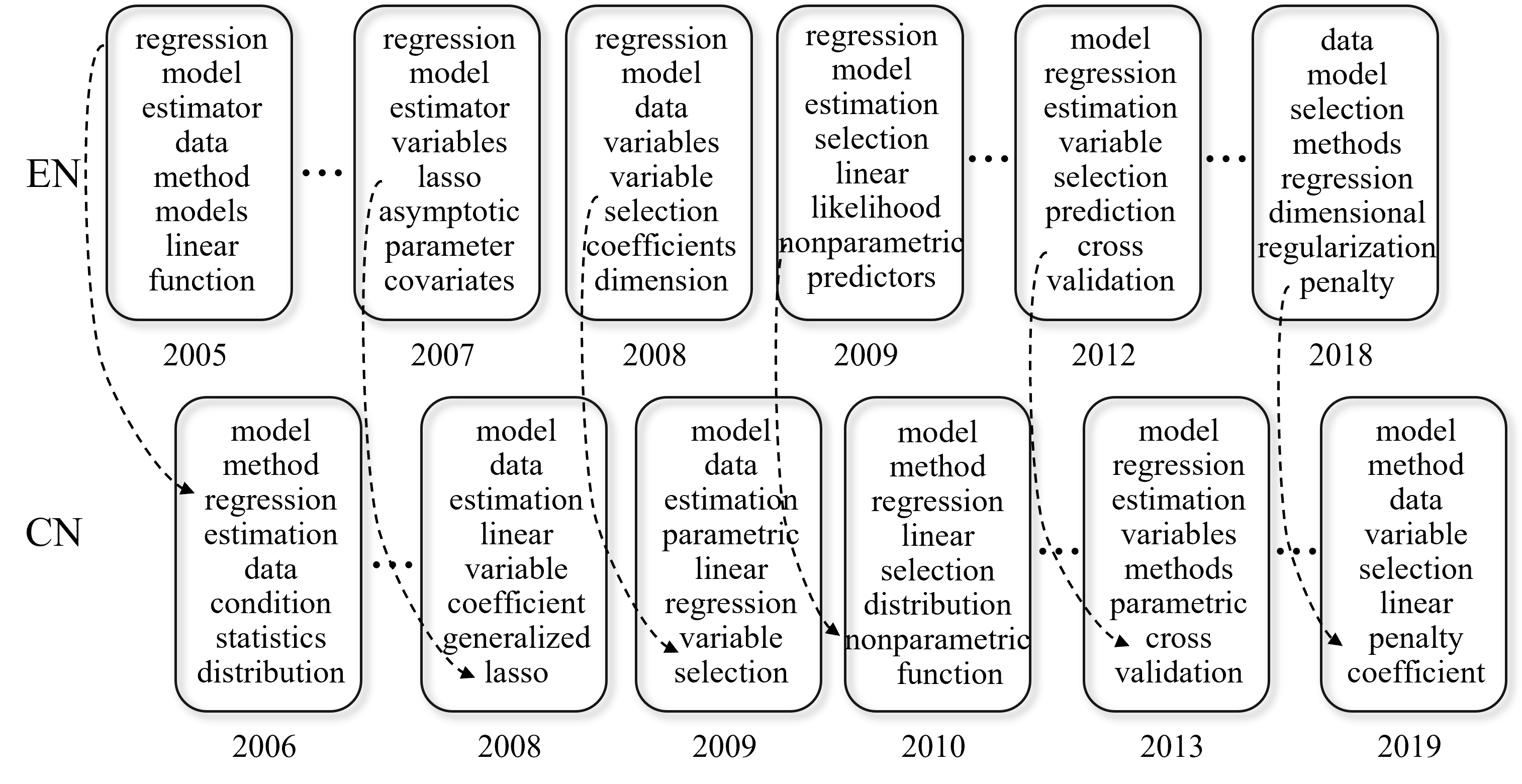}
	}
	\caption{Evolution of eight most commonly used words for two shared topics in the leading corpus (EN) and lagged corpus (CN). The dotted line implies the $l=1$ time period lead-lag relationship in two corpora. It is obvious that, the commonly used words in the leading corpus at time $t$ are very similar with those used in the lagged corpus at time $t+1$. This implies the existence of the lead-lag relationship between the two corpora. }
	\label{fig:lead_lag_relation}
\end{figure}

Next, we examine the lead-lag relationship discovered by our model. For illustration purpose, we randomly select two shared topics estimated by the JDETM with the lag time period $l=1$. Then in each time period, we select documents having the largest proportions in this topic from both corpora. Next, we find the most commonly used words in these documents. Figure \ref{fig:lead_lag_relation} shows the evolution of eight most commonly used words in the selected documents in the leading corpus and lagged corpus. As shown, the commonly used words in the leading corpus at time $t$ are very similar with those used in the lagged corpus at time $t+1$. Furthermore, the evolution of words in the shared topic ``Model Selection'' shows the temporal changes in academic hotspots: from the linear regression models in early years, to various variable selection methods (e.g., Lasso), and more recently nonparametric methods and general discussions about regularization and penalty. These results imply the content influences of the leading corpus made on the lagged corpus. It further verifies the existence of the lead-lag relationship between the two corpora.

Finally, we discuss the forecast ability of our proposed method. The lead-lag relationship can help predict the topics to be discussed by the lagged corpus in the future. Specifically, the lead-lag relationship in JDETM is represented by the \emph{shared topics} with an $l$-period time interval between the leading corpus and lagged corpus. As a consequence, the shared topics discussed by the leading corpus at the last time slice $T$ should also be discussed by the lagged corpus at the future time slice $T+l$. By this way, the JDETM can predict the future topics to be discussed by the lagged corpus. In Table \ref{tab:shared_topics}, we provide the ten shared topics discussed by the English journal papers in the last year 2019. For each topic, the top seven words with the highest probabilities are reported. As suggested by JDETM, these topics would be talked about by the Chinese thesis in the year 2020.
\begin{table}[thp]
	\caption{The top seven words with the highest probabilities under ten shared topics in the last year 2019. }
\renewcommand\arraystretch{1.2}
\footnotesize
	\label{tab:shared_topics}
	\begin{tabular}{cll}
    \hline
    &Shared Topic &Top Seven Words \\
    \hline
    1&Economic Development	&economic, development, industry, energy, efficiency, carbon, environmental\\
	2&Practical Application &method, problem, study, existing, practical, theoretical, comparison\\
    3&Parameter Estimation	&estimation, likelihood, parameter, maximum, empirical, Bayesian, copula\\
    4&Time Series Analysis	&time, series, predict, spatial, dynamic, autoregressive, nonlinear\\
    5&Financial Markets	&market, stock, financial, volatility, return, portfolio, Garch\\
    6&Risk Management	&financial, price, rate, credit, risk, exchange, banks\\
    7&Stochastic Process	&random, stochastic, process, probability, differential, convergence, markov\\
    8&Model Selection	&regression, variable, selection, nonparametric, dimensional, quantile, Lasso\\
    9&Data Analysis	&data, analysis, statistical, information, methods, statistics, principal\\
    10&Experiment Design	&design, optimal, system, control, decision, uncertainty, controller\\
	\hline
	\end{tabular}
\end{table}

\clearpage
\section{Conclusion and Discussion}

In this work, we focus on the topic modeling problem of two text corpora, among which one leading corpus could have influence on the topics to be discussed by the lagged corpus in the future. We define this content influence as the lead-lag relationship between the two corpora. To recognize this relationship, we propose a jointly dynamic topic modeling approach. By assuming three types of topics, i.e., the shared topics, the lead-specific topics and the lag-specific topics, the jointly dynamic topic modeling approach can not only discover the specific patterns in each corpus, but also find out their similarities characterized by the shared topics. To handle large-scale textual datasets, an embedding extension of the jointly dynamic topic model is developed. The finite sample performances of the proposed models are numerically investigated by experiments on synthetic data. Finally, we apply the jointly dynamic embedding topic model on two statistical text corpora consisting of published papers and graduation theses to show the application of the proposed model.

To conclude this work, we consider several directions for future study. First, the lagged time period $l$ should be specified in advance. Then how to select an optimal $l$ is worth of consideration. Second, to avoid setting the numbers of three types of topics, they can be further modeled using the hierarchical Dirichlet process. The numbers of three types of topics should also allow to change over time. Third, the lead-lag relationship allows to forecast the shared topics to be discussed by the lagged corpus in the future. However, how to forecast the popularity of these shared topics (i.e., the $\phi_d'$s) in the future needs further consideration. Last, the proposed jointly dynamic topic modeling approach can be further combined with other dynamic topic models to recognize the lead-lag relationship in more complex situations.

\section*{Acknowledgement}
This work is supported by National Natural Science Foundation of China (72001205, 11971504), fund for building world-class universities (disciplines) of Renmin University of China, the Fundamental Research Funds for the Central Universities and the Research Funds of Renmin University of China (2021030047), Foundation from Ministry of Education of China (20JZD023), Ministry of Education Focus on Humanities and Social Science Research Base (Major Research Plan 17JJD910001).

\section*{Appendix: The Lists of Selected Journals and Top Schools with the Highest Number of Theses}
\spacing{1}
    \label{app:JorList}
        \begin{table}[h]
        \small
        \begin{center}
        \renewcommand\arraystretch{1.3}
        \begin{tabular}{cl}
            \hline
            \multicolumn{2}{c}{The Ten Selected Statistical Journals}\\
            \hline
            1 & Journal of the American Statistical Association \\
            2 & Econometrica  \\
            3 & Journal of the Royal Statistical Society Series B (Statistical Methodology) \\
            4 & Annals of Statistics \\
            5 & Fuzzy Sets and Systems \\
            6 & Computational Statistics \& Data Analysis \\
            7 & American Statistician \\
            8 & Journal of business \& Economic Statistics \\
            9 & Stochastic Processes and Their Applications  \\
            10 & Statistics and Computing \\
            \hline
            \multicolumn{2}{c}{The Chinese Universities with Top Number of Theses}\\
            \hline
           1 & Dongbei University of Finance and Economics \\
           2 & Zhejiang Gongshang University  \\
           3 & East China Normal University    \\
           4 & Huazhong University of Science and Technology\\
           5 & Tianjin University of Finance and Economics\\
           6 &  Northeast Normal University\\
           7 & Hunan University\\
           8 &   Jinan University\\
           9 & Central South University\\
           10 & Shandong University\\
            \hline
        \end{tabular}
        \end{center}
    \end{table}

\newpage
\bibliographystyle{spbasic}
\bibliography{reference}
\end{document}